\documentclass{article}

\usepackage[preprint]{neurips_2025}

\usepackage[utf8]{inputenc} % allow utf-8 input
\usepackage[T1]{fontenc}    % use 8-bit T1 fonts
\usepackage{hyperref}       % hyperlinks
\usepackage{url}            % simple URL typesetting
\usepackage{booktabs}       % professional-quality tables
\usepackage{amsfonts}       % blackboard math symbols
\usepackage{nicefrac}       % compact symbols for 1/2, etc.
\usepackage{microtype}      % microtypography
\usepackage{xcolor}         % colors

\usepackage{adjustbox} % 必需: 支持表格调整和缩放
\usepackage{booktabs} % 必需: 高级表格排版
\usepackage{arydshln} % 必需: 支持虚线
\usepackage{multicol}
\usepackage{multirow}
\usepackage{amsmath}
\usepackage{makecell}
\usepackage{wrapfig}
\usepackage{tcolorbox}

\title{UItron: Foundational GUI Agent with Advanced Perception and Planning}

\author{Zhixiong Zeng$^{\dag}$, Jing Huang$^{\dag}$, Liming Zheng, Wenkang Han, \\ 
\textbf{Yufeng Zhong, Lei Chen, Longrong Yang, Yingjie Chu, Yuzhi He, Lin Ma$^{\ast}$} \\
Meituan \\
\texttt{zengzhixiong@meituan.com, forest.linma@gmail.com} \\
Project: \url{https://github.com/UITron-hub/UItron}
}

\begin{document}
{\let\thefootnote\relax\footnotetext{$^{\dag}$ Equal contribution. $^{\ast}$ Corresponding author.}}

\maketitle

\begin{figure}[!h]
    \centering
    \vspace{-15pt}
    \includegraphics[width=0.9\linewidth]{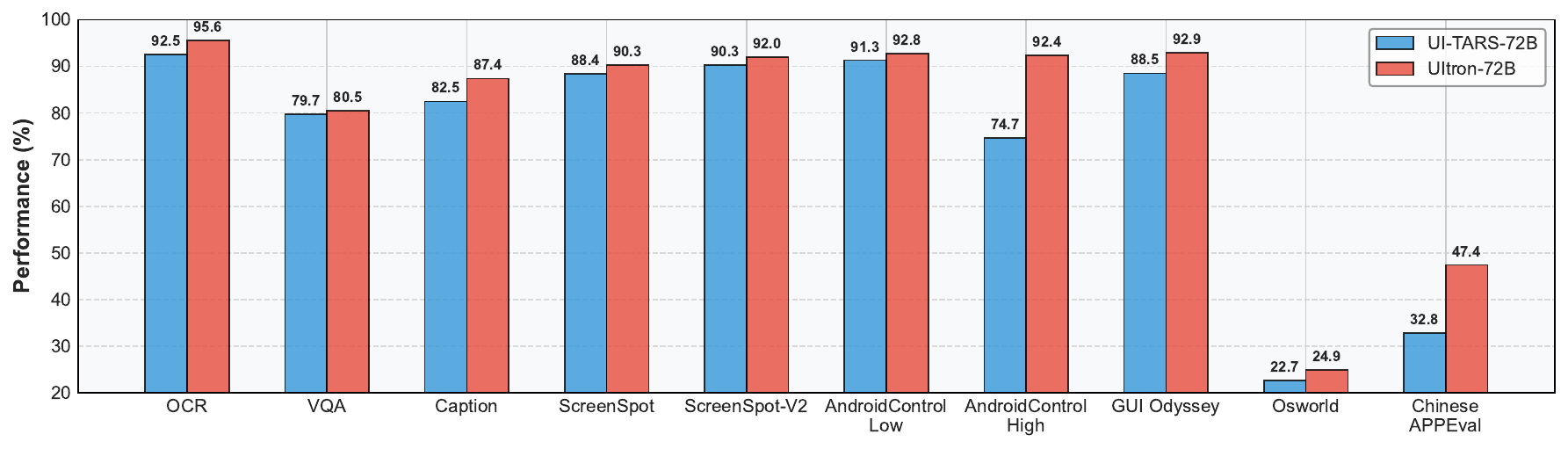}
    \vspace{-10pt}
    \caption{Comparison of UItron and UI-Tars in perception/grounding/planning and Chinese scenarios.}
    \label{fig:fig1}
    % \vspace{-5pt}
\end{figure}

\begin{abstract}
GUI agent aims to enable automated operations on Mobile/PC devices, which is an important task toward achieving artificial general intelligence. The rapid advancement of VLMs accelerates the development of GUI agents, owing to their powerful capabilities in visual understanding and task planning. 
However, building a GUI agent remains a challenging task due to the scarcity of operation trajectories, the availability of interactive infrastructure, and the limitation of initial capabilities in foundation models.
In this work, we introduce UItron, an open-source foundational model for automatic GUI agents, featuring advanced GUI perception, grounding and planning capabilities.
UItron highlights the necessity of systemic data engineering and interactive infrastructure as foundational components for advancing GUI agent development. 
It not only systematically studies a series of data engineering strategies to enhance training effects, but also establishes an interactive environment connecting both Mobile and PC devices.
In training, UItron adopts supervised finetuning over perception and planning tasks in various GUI scenarios, and then develop a curriculum reinforcement learning framework to enable complex reasoning and exploration for online environments.
As a result, UItron achieves superior performance in benchmarks of GUI perception, grounding, and planning.
In particular, UItron highlights the interaction proficiency with top-tier Chinese mobile APPs, as we identified a general lack of Chinese capabilities even in state-of-the-art solutions.
To this end, we manually collect over one million steps of operation trajectories across the top 100 most popular apps, and build the offline and online agent evaluation environments. Experimental results demonstrate that UItron achieves significant progress in Chinese app scenarios, propelling GUI agents one step closer to real-world application.

\end{abstract}

\section{Introduction}

GUI agents~\cite{hong2024cogagent,zhang2024you,zhang2024android,yang-etal-2025-aria,wuatlas,gou2025uground,xu2025aguvis,qin2025ui,lin2025showui,chen2025less} aim to automatically execute complex tasks in various digital environments such as PC and Mobile, satisfying the growing expectations of autonomous decision-making and software control in human-computer interaction.
These agents decompose the task instructions into multi-step actions by observing the screen status, then navigate and manipulate the on-screen elements following the human-like interactive manners (\emph{i.e.}, click, scroll).
This human-like approach provides visually trackable trajectories with step-by-step task execution process, enabling convenient user interaction and explainable decision-making.
Therefore, GUI agents have received a rapidly growing amount of attention, becoming an important research topic toward achieving artificial general intelligence.

%GUI智能体在很远的早期。。。--》得益于LLM涌现的推理能力，智能体可以推理和规划--》早期的工作以html...--〉随着mllm的发展...

The pursuit of automated GUI agent has been going on for a long time.
Early methods~\cite{deng2023mind2web,gur2024real,lai2024autowebglm} utilize optical character recognition and icon detectors to parse GUI environments into textual elements (e.g., HTML and AXTree) as input to the LLM, leveraging its powerful reasoning capabilities to plan and generate multi-step executable actions.
The rapid advancement of vision-language models catalyze a series of GUI agents (e.g.,~\cite{yang-etal-2025-aria,wuatlas,gou2025uground,xu2025aguvis,qin2025ui}) that operate directly on visual GUI images, which achieve superior performance within the framework of unified perception and planning in pure vision.
A representative work is UI-TARS~\cite{qin2025ui}, which achieves leading performance via a large amount of data engineering and a carefully designed iterative training framework.
Recently, some R1-style works~\cite{lu2025ui,liu2025infigui,zhou2025gui,tang2025gui,yang2025zerogui,dong2025agentic} represented by GUI-R1~\cite{zhou2025gui} designs multimodal reasoning data and explores the typical group relative policy optimization to improve reasoning ability, reporting improved results in grounding benchmarks.
To address the limited adaptability in offline environments, ZeroGUI~\cite{yang2025zerogui} and ARPO~\cite{dong2025agentic} propose the online reinforcement learning framework, which adopts VLM-based automatic reward estimation to assess task success and continuously learn from the GUI environments, without hand-crafted evaluation functions.

However, developing automatic GUI agents still remains a highly challenging task due to several limitations: the scarcity of annotated operation trajectories, the availability of interactive infrastructure, and the limitation of initial capabilities in foundation models.
As illustrated in Figure 1, training a GUI agent necessitates precise GUI perception, grounding, offline planning, and online planning capabilities, thereby rendering the collection of adequately annotated trajectory data exceedingly difficult. Moreover, in contrast to traditional multimodal tasks, GUI agents face the additional challenge of requiring an interactive environment to execute the model-generated actions that enables multi-round interaction. Additionally, extensive empirical evidence shows that current foundation models typically possess limited performance in GUI scenarios, substantially hindering progress toward effective GUI agent development.

In this paper, we introduce UItron, a powerful open-source foundational model for automatic GUI agents, with powerful GUI perception, task grounding, offline/online planning capabilities.
UItron emphasizes the importance of data engineering and interactive infrastructure for developing GUI agents. For data engineering, we significantly expand the available operation trajectories through three key aspects: data unification, trajectory distillation, and manual annotation over different domains. 
Moreover, we systematically investigated a series of data engineering strategies to enhance training effectiveness, including the utilization of various trajectory elements (\emph{e.g.}, observation, thought and action), the exploration of different reasoning formats, and the incorporation of diverse reflection mechanisms like backtracing.
We also find the advantages of integrating multi-task UI-related data and general multimodal data. 
For interactive infrastructure, we build an interactive environment connecting both Mobile and PC devices. It not only simplifies trajectory data collection by automatically recording screenshots and coordinates, but also provides a foundation for online reinforcement learning (RL) during training.

During training, we employ a three-stage training strategy over several GUI scenarios, which includes GUI perception, planning and RL. Note that the RL stage is specifically designed to enhance complex reasoning and exploration capabilities within online environments.
First, UItron adopts a supervised finetuning strategy for GUI perception and planning tasks. 
The perception task focuses on improving the basic understanding ability of the vision-language model in GUI scenarios, such as grounding, captioning, VQA, and OCR. The planning stage concentrates on forecasting the next action based on historical actions.
Then UItron develops a Curriculum Reinforcement Learning (CuRL) framework with group relative policy optimization algorithm on trajectory data.
To address the problem of sparse rewards, CuRL first computes dense rewards from the action steps in the offline environment (simple), and then computes the task-level reward for the trajectory data in the online environment (complex).
In addition, to improve the credibility of rewards in the RL process, we strictly filter the trajectories that are predicted correctly by multiple scoring models simultaneously.

In particular, UItron emphasizes its ability to interact with top-tier mobile Apps in China, as we find that even state-of-the-art solutions generally underperform in Chinese scenarios for GUI agents.
To this end, we meticulously annotate over one million action steps from the top 100 monthly active Apps, covering mainstream interaction scenarios such as social networking, office work, entertainment, and shopping. Based on this, we constructed an offline evaluation dataset to assess the capabilities of GUI agents in Chinese App scenarios, evaluating the performance of different models based on two classical evaluation metrics: single-step success rate and task completion rate. 
To evaluate the realistic interaction performance of GUI agent in real-world applications, we also build an Android-based cloud real-device environment for online evaluation. Specifically, we develop a rollout method to alternately execute actions and refresh status between GUI agent and Android-based cloud environment.
Next, we developed an automated scoring mechanism that leverages multiple VLMs to score the entire task trajectory, and averages the scores to generate evaluation results. 
Experimental results not only confirm the limitations of existing methods, but more importantly, demonstrate that UItron has achieved substantial progress in Chinese application scenarios, advancing GUI agents toward practical and real-world deployment.

The main contributions of this work are summarized as follows:
\begin{itemize}
    \item We present a systematic investigation of data engineering and interactive infrastructure that effectively supports the development of foundational GUI agents.
    \item We develop a curriculum reinforcement learning framework with dense and credible rewards for trajectory data in GUI agents.
    \item We open-source UItron, achieving superior performance in benchmarks of GUI perception, grounding and offline planning, and competitive results in online agent environments.
    \item We significantly improves the interactive capabilities of UItron in Chinese scenarios through carefully labeled data and tailored online environments.
\end{itemize}

%Early methods implement rule-based or script-based GUI agents to manage various GUI operations, and subsequent approaches using deep learning to improve GUI content understanding.

\begin{figure}[t]
    \centering
    \includegraphics[width=0.85\linewidth]{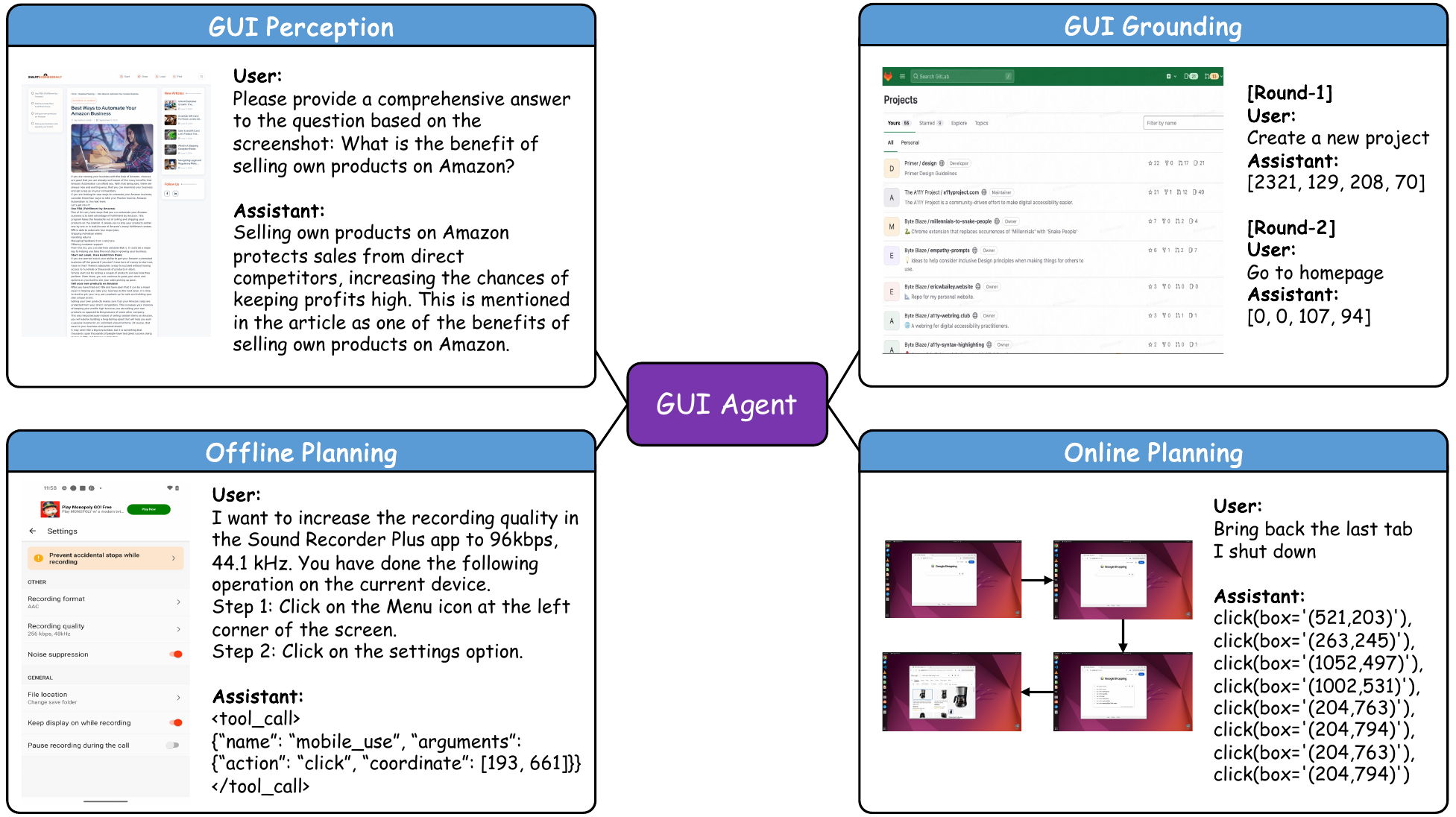}
    \caption{The core capabilities of GUI agent, including GUI perception, grounding, offline planning and online planning.}
    \label{fig:fig3}
    \vspace{-5pt}
\end{figure}

\section{Related Works}
\subsection{MLLM}
% Visual encoder and projector
Large language models (LLMs) have shown strong generalization capabilities and instruction-following abilities. However, they can only process text information, while real-world applications require models to understand visual information. Thus, Multimodal Large Language Models (MLLMs) such as GPT-4~\cite{achiam2023gpt} and LLaVA~\cite{liu2023visual} utilize visual encoders and visual projectors to integrate visual data into large language models.
MLLM works on visual encoder mainly focus on improving input resolution, which can be roughly divided into direct scaling and patch-division. Direct scaling, \textit{e.g.},~\cite{bai2023qwen,liu2024improved,hong2024cogagent}, inputs higher-resolution images into the encoder, usually requiring further fine-tuning of the encoder or the use of a pre-trained encoder with higher resolution. Patch-division, \textit{e.g.},~\cite{lin2023sphinx,li2024monkey}, splits high-resolution images into multiple patches and then reuses the low-resolution encoder, which has gradually become the mainstream choice because it can support dynamic resolution or natural resolution~\cite{chen2024far,wang2024qwen2,wu2024deepseek,bai2025qwen2,guo2025seed1}, making it suitable for processing varying visual information.
MLLM works on projector mainly focuses on the effective mapping of visual information, which can be roughly divided into token-level and feature-level. The token-level projector converts the output features into tokens and concatenates them with text tokens before feeding them into the large language model. Some works use Q-Former, \textit{e.g.},~\cite{zhang2023video,dai2023instructblip}, but most works directly use MLP to bridge the modal gap due its simplicity and generalization, \textit{e.g.},~\cite{liu2023visual,su2023pandagpt,liu2024improved,wang2024qwen2,bai2025qwen2}.
The feature-level projector, \textit{e.g.},~\cite{alayrac2022flamingo,zhang2023llama,bai2023qwen,wang2024cogvlm}, inserts additional modules that enable in-depth interaction and fusion between text and visual features. For example, Flamingo~\cite{alayrac2022flamingo} inserts additional cross-attention layers into the frozen LLM layers to enhance language features using external visual cues.

% Data
Beyond architecture, scaling data is key to improve the performance of MLLMs, and numerous studies have focused on enhancing the data quality for MLLMs.
For example, many MLLM studies focus on improving the data quality of instruction fine-tuning, \textit{e.g.},~\cite{dai2023instructblip,zhang2023llama,wang2023visionllm,gao2023llama,luo2023cheap,chen2024internvl,liu2024llavanext,chen2024sharegpt4v,guo2025seed1}. 
In addition, some studies, \textit{e.g.},~\cite{wang2022self,liu2023visual,zhu2023minigpt,yang2023gpt4tools,wang2023see,chen2024allava}, collect samples through self-instruction, where they use large language models to generate data that conform to text instructions based on a small number of manually annotated samples. 
Then, language data is usually mixed with multimodal data to train the model~\cite{ye2023mplug,luo2023cheap,gao2023llama}, improving the instruction-following ability. 
In addition, preference alignment is often used in scenarios where the model needs to align with specific human preferences. 
Reinforcement Learning from Human Feedback (RLHF)~\cite{ouyang2022training} and Direct Preference Optimization (DPO)~\cite{rafailov2023direct} are two main techniques for preference alignment.

\subsection{GUI Agent}
% HTML or AXTree
Early GUI agents rely on HTML or AXTree data to describe the screen state of user interactions through textual state descriptions.
These methods depend on the structured representation of web page elements, enabling agents to locate targets based on tags, attributes, or text content.
Raw HTML or AXTree data usually has redundant or noisy structures.
Thus, some methods have studied the extraction of effective information from HTML. 
For example, Mind2Web~\cite{deng2023mind2web} uses a fine-tuned language model to sort web page elements and extract important ones. WebAgent~\cite{gur2024real} uses a dedicated HTML-T5 model to generate HTML fragments for specific tasks. AutoWebGLM~\cite{lai2024autowebglm} designs an algorithm to simplify HTML content.
Some methods~\cite{he2024webvoyager,yang2023setofmark} combine visual information. For instance, Set-of-Mark~\cite{yang2023setofmark} integrates visual and tree tag information for agent decision-making.
However, the quality of HTML or AXTree data can limit the application of the aforementioned methods, considering different standards across platforms.
In addition, structured data requires meticulous pre-processing, and in some cases, it also relies on additional heuristic methods or trained models to accurately locate and understand key GUI components.

% MLLM
With the rise of multimodal large language models (MLLMs), an increasing number of pure vision-based methods have been proposed.
These methods leverage the strong visual capabilities of multimodal language large models and eliminate the need for manually designing data pre-processing for each task, so they have significant advantages over early GUI agents in terms of generalization.
Early pure visual GUI agents focus on using MLLMs to process screenshots to understand GUI components, replacing HTML-based GUI component understanding. 
For example, CogAgent~\cite{hong2024cogagent} uses MLLMs to process high-resolution GUIs, achieving performance that surpasses HTML-based methods. Auto-GUI~\cite{zhang2024you} unifies GUI grounding into a text-driven grounding task and proposes action chains, using a series of historical actions to enhance agents; COAT~\cite{zhang2024android} further models the thinking process of "what action should be performed" to improve agents in complex tasks.

% MLLM - data
Furthermore, researchers have found that scaling data is key to enhancing the performance of GUI agents, thus proposing using synthetic data or video data.
For instance, Aria-UI~\cite{yang-etal-2025-aria} proposes an extensible data synthesis pipeline for generating grounding data, which is used to train MLLMs specialized in GUI grounding. OS-Genesis~\cite{yang-etal-2025-aria} proposes a trajectory synthesis method to retrospectively generate tasks from the agent-environment interaction, rather than relying on manual supervision or predefined tasks. GUI-explorer~\cite{xie2025gui} automatically generates function-aware task objectives by analyzing GUI structure information, and achieves low-cost generalization of agents through unsupervised analysis of state transitions in the observation-action-result triples. OS-Atlas~\cite{wuatlas} releases the first multi-platform GUI data synthesis toolkit, supporting the automatic synthesis of cross-platform GUI grounding data while resolving action naming conflicts. GUI-Xplore~\cite{sun2025gui} enables GUI agents to learn from exploration videos.

% MLLM - human-like agents
Recently, researchers have aimed to make GUI agents more similar to human-like agents. For example, UGround~\cite{gou2025uground} executes actions only through human-like keyboard and mouse operations. Aguvis~\cite{xu2025aguvis} unifies different GUI action spaces and divides training into grounding and planning, enhancing the action ability after the agent has high GUI grounding performance. GUI-Odyssey~\cite{lu2025gui} presents a large-scale cross-application GUI agent training and evaluation dataset, allowing agents to interleavingly use multiple applications to execute tasks. UI-TARS~\cite{qin2025ui} can perform human-like interactions, which achieves accurate perception of GUI elements by collecting a large amount of screenshot data and enhances agent capabilities through various reasoning modes.

% MLLM - efficiency
The efficiency of GUI agents is crucial for practical deployment, so some research efforts are also dedicated to this. For example, ShowUI~\cite{lin2025showui} constructs a UI correlation graph to identify redundant UI relationships and selects tokens based on the identification results to improve training efficiency. SimpleAgent~\cite{chen2025less} masks redundant elements in the current environment and uses consistency constraints to guide the cropping of historical tokens.

% RL
However, the aforementioned methods mainly focus on using Supervised Fine-Tuning (SFT) to fit manually annotated action trajectories for enhancing the performance of GUI Agents. 
In fact, numerous studies have shown that the single SFT method limits the model ability for autonomous exploration, which is crucial in GUI Agent tasks.
Early works~\cite{luo2025gui,lu2025ui} directly optimize models using a 0-1 reward similar to deepseek-R1~\cite{guo2025deepseek}. 
For example, UI-R1~\cite{lu2025ui} uses rule-based action rewards and the GRPO algorithm to enhance the performance of GUI agents in unknown scenarios.
Further optimizations have focused on issues such as reasoning patterns~\cite{liu2025infigui,zhou2025gui}, sparse rewards~\cite{yuan2025enhancing,tang2025gui}, and online learning~\cite{tang2025gui,wei2025webagent}.
InfiGUI-R1~\cite{liu2025infigui} proposes to first use trajectories with explicit reasoning steps for training, and then apply reinforcement learning to enhance the error correction ability.
GUI-G1~\cite{zhou2025gui} uses a fast-thinking template to encourage the model to generate answers directly, reducing excessive reasoning during training, and simultaneously designs a difficulty-aware RL objective to better learn hard samples.
SE-GUI~\cite{yuan2025enhancing} calculates continuous rewards using the proximity between the predicted box and the GT box, replacing the 0-1 reward to alleviate the sparse reward problem.
GUI-G$^2$~\cite{tang2025gui} transforms the discrete classification of GUI grounding into the continuous optimization of IoU through Gaussian point rewards, coverage rewards, and an adaptive variance mechanism, addressing the sparse reward.
Zero-GUI~\cite{yang2025zerogui} proposes the ZeroGUI online learning framework, which automatically generates tasks and estimates rewards.

\subsection{API Agent}
Another type of agent, distinguished by its mode of interaction with computers or mobile phones, is the API-centric agent. These agents interact with external tools, functions, or services through pre-defined, well-structured programming interfaces.
During interaction, relevant API information (\textit{e.g.}, function names) is included in the LLM prompt. The agent receives natural language requests from users and selects the most appropriate API to execute the task.

Microsoft Copilot is a typical example of an enterprise-level API agent. Through interfaces such as the Microsoft 365 Copilot API, it allows developers to integrate capabilities like data analysis and document generation into custom applications.
Related research focuses on issues such as automated generation/updating of tools~\cite{gao2024clova,wang2024trove,trivedi2024appworld,wang2024llms,wang2024llms}, simplification of tools~\cite{yuan2025easytool}, and patterns of tool usage~\cite{du2024anytool,lin2025robust}.
CLOVA~\cite{gao2024clova} identifies tools that need updating by analyzing human feedback, automatically collects training data, and uses prompt tuning to update the tools.
TroVE~\cite{wang2024trove} constructs a verifiable and efficient function toolbox through generating, using, expanding, and periodically streamlining the toolbox.
Appworld~\cite{trivedi2024appworld} builds a high-quality execution environment for agents and creates a set of autonomous agent task sets that require agents to generate cross-application interaction code for processing.
STE~\cite{wang2024llms} utilizes large models to simulate reasonable environments for tool usage, then enables large models to interact with tools and learn from environment feedback.
Kimi K2~\cite{team2025kimi} proposes a trajectory synthesis scheme for function calls, relying on a vast tool specification library constructed from real-world tools and synthesized tools.
Then, EasyTool~\cite{yuan2025easytool} converts tool documents into unified and concise instructions to improve tool usage efficiency.
Anytool~\cite{du2024anytool} retrieves APIs to handle user needs and proposes a self-reflection mechanism.
Hammer 2.1~\cite{lin2025robust} improves the model sensitivity to irrelevant functions through enhanced datasets and function masking technology.
API agents rely on text-based API calls, which are generally highly reliable, and can complete complex tasks with a single call. However, they are limited to pre-defined APIs, have low transparency, and lack the generalization of human-like actions exhibited by GUI agents.

\section{UItron}
\label{method}
% 三大能力：perception、grounding、offline planning、online、中文
% 数据工程：data cleaning, trajectory distillation, and manual annotation，整合非轨迹和一般多模态数据
% 交互式基建：自动保存截图和坐标来收集数据，简化数据流程，提供RL环境。采用offline single-step reward and online multi-step reward
% 中文app能力

UItron is an open-source foundational GUI agent framework designed to advance automated interaction and reasoning across both mobile and PC environments. The system is built upon two key pillars: a robust data engineering pipeline tailored for GUI agent training, and a unified interactive infrastructure that supports scalable data collection and dynamic training. Leveraging these foundations, UItron delivers core capabilities in perception, grounding, and planning, enabling agents to understand complex interfaces, accurately localize tasks, and execute effective action sequences in diverse real-world scenarios.

\subsection{Problem Formulation}

GUI Agent aims to predict the next action in the $n$-th step based on the task instruction, historical actions and visual environment observation (a GUI image). The action is usually represented by the action type and parameters, such as the click(box) and input(content).
Formally speaking, we denote the task instruction as $T$, the historical action as $\{a_1,a_2,...,a_{n-1}\}$, and the visual environment observation as $o_n$. Therefore, the task goal of GUI agent in the $n$-th step can be formulated as:
\begin{equation}\label{eq_Future}
    a_n = \mathbf{M}_\theta(T, (a_1, a_2,...,a_{n-1}),o_n),
\end{equation}
here $\mathbf{M}$ represents the GUI agent with trainable paratemers $\theta$.

Note that previous work in this area usually utilizes multiple historical images to augment the input with historical information. It significantly increases the length of the input sequence and the computational cost, which is detrimental to the redundant nature of visual information. In fact, we empirically found that omitting historical images does not result in a significant performance degradation in most benchmarks, as historical action information provides sufficient gains. Therefore, to mitigate computational cost, we did not use historical images in this version.

\subsection{Data Engineering}
% A robust and generalizable GUI agent requires not only large-scale data, but also high-quality, diverse, and well-structured datasets that accurately reflect real-world application scenarios. To meet these requirements, we have developed a comprehensive data engineering pipeline that systematically addresses the limitations of existing resources through four key components: data cleaning and standardization, trajectory distillation, manual annotation, and multimodal integration.
As shown in Figure 2, we explore systematic data engineering to improve UItron, including perception data, planning data, and distillation data.
Besides, we also organize a small amount of general multimodal data that is beneficial to GUI agent, as well as high-quality manual annotation data for Chinese scenarios.

\begin{figure}[t]
    \centering
    \includegraphics[width=1.0\linewidth]{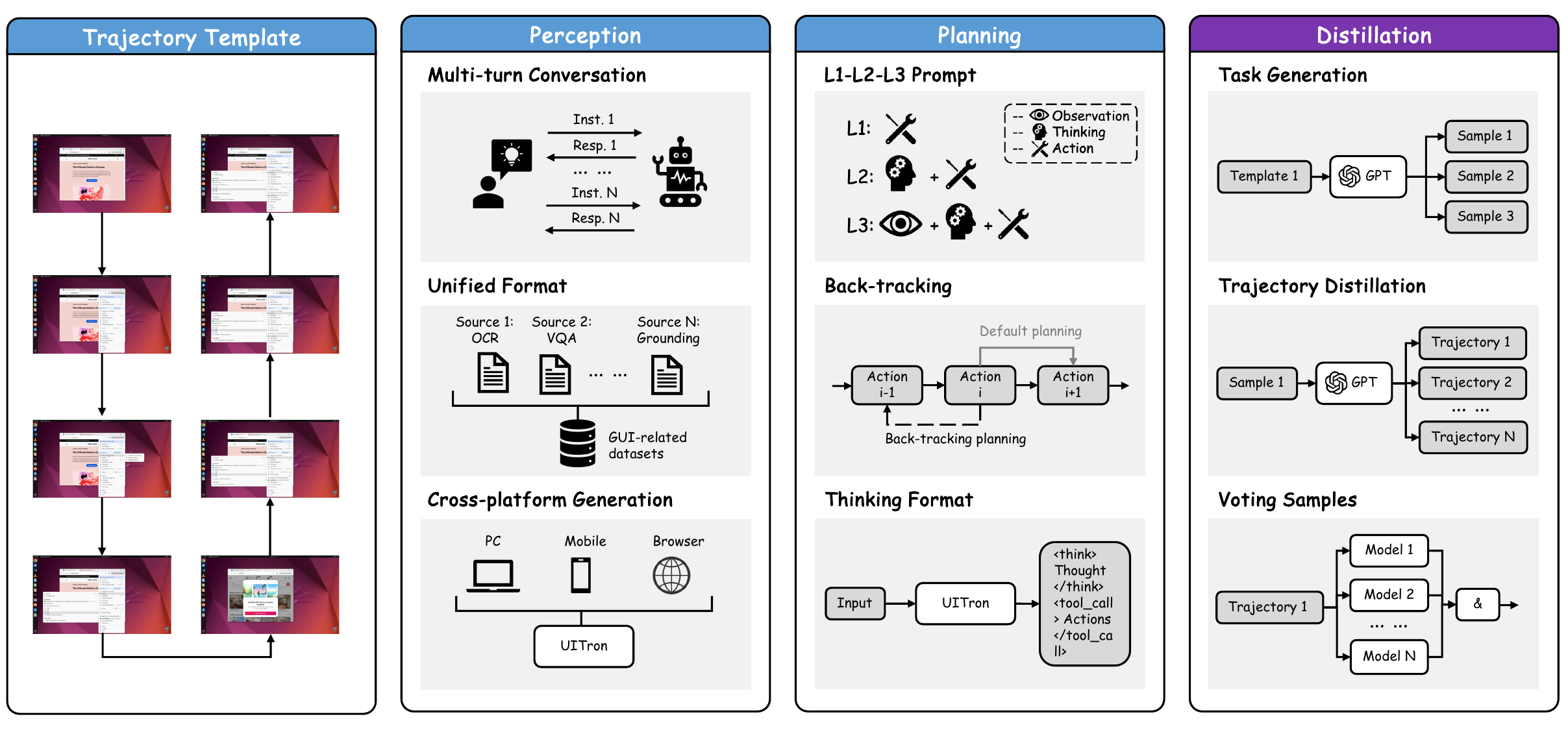}
    \vspace{-15pt}
    \caption{Overall introduction of data engineering.}
    \label{fig:fig2}
    \vspace{-5pt}
\end{figure}

\subsubsection{Perception Data}
\noindent \textbf{Multi-turn Conversation. }
In practical applications, a single complex screenshot can contain hundreds of UI elements, and open-source grounding datasets typically feature multiple objects within one image. To minimize redundant image loading and decrease training costs, we consolidated various instruction/description-answer pairs associated with the same screenshot into unified multi-turn conversations, thereby constructing multi-turn training samples. Utilizing such multi-turn data for training not only lowers computational overhead but also improves the model’s ability to comprehend and distinguish between different elements within a UI scene.

\noindent \textbf{Multi-task Unification.} 
To enhance the basic understanding ability in GUI scenarios, we collect a large amount of UI-related perception data instead of just considering the traditional agentic trajectory data. We collect a wealth of image-text multimodal pairs from a wide range of PC/mobile application screenshots, covering tasks in GUI scenarios such as OCR, VQA, and Caption. We then integrate these UI-related perception data and traditional agentic trajectory data into the unified format to support training.

\noindent \textbf{Cross-platform Generalization.} 
% 我们的方法统一了开源数据集（包括a,b,c），系统地研究了不同的合成标准是否可以相互补充，从而提高了智能体定位在不同场景下的泛化能力。
Although a substantial amount of grounding data already exists in the GUI agents field, but data collected from different platforms and devices often lacks generalizability, and various tasks employ distinct and isolated data synthesis criteria, making it challenging for these datasets to complement one another.
To address the generalization challenge in GUI grounding, we integrated data from diverse sources and synthesis methodologies within the GUI agent domain. By unifying open-source datasets (including Uground~\cite{gou2024navigating}, Aria-UI~\cite{yang2024aria}, Aguvis~\cite{xu2024aguvis} and OS-Atlas\cite{wu2024atlas}), our approach systematically explores whether diverse synthesis criteria can complement one another, thereby enhancing the generalization capability of agent localization across various scenarios.

\subsubsection{Planning Data}
\noindent \textbf{L1-L2-L3 Inference.} 
% 一个planning任务的执行除了最终要输出的动作以外，还可以通过多种层次的感知和思考辅助动作的预测。参考之前的工作，我们将planning阶段的感知分为屏幕观察、思考、动作和总结几个层次。从最简单的l1只预测动作和简单总结，到l2在l1的基础上增加了思考，再到添加屏幕上下文观察分析UI界面的变化，这种多层细混合的感知方式能够帮助模型适应不同难度、不同场景下的任务。为了平衡效率和准确性，我们在推理时使用l2层次的描述作为历史上下文提示模型进行动作预测。
In addition to the final output action, the execution of a planning task can be enhanced by incorporating multiple levels of perception and reasoning to facilitate action prediction. Following ~\cite{xu2025aguvis}, we divide the planning data into several elements including screen observation, reasoning (thinking), action and summarization. 
The L-1 inference involves only action prediction and summarization, L-2 inference further introduces reasoning, and L-3 inference incorporate screen context to observe and analyze changes in the UI interface. 
This multi-layered and fine-grained perception strategy enables the model to better adapt to tasks of varying complexity and diverse scenarios. To balance efficiency and accuracy, we utilize L2-level descriptions as historical context prompts for action prediction during inference.

\noindent \textbf{Back-tracking.} 
% GUI智能体的规划过程可以天然地被建模为部分可观察的马尔可夫决策，模型根据历史的动作和当前状态预测下一个动作。
The planning process of a GUI agent can be naturally formulated as a partially observable Markov decision process, in which the model predicts the next action based on historical actions and the current state.  However, this approach neglects the model’s capacity for reflection and backtracking on previous decisions.  Specifically, while the model is aware of its current state, it lacks insight into the sequence of actions that led to that state. Consequently, the model struggles to establish connections between past, present, and future states, which hinders its ability to generate consistent and coherent action predictions. Following \cite{huang2025scaletrack}, we enhance the interaction between GUI agents and their environment by introducing backtracking. Specifically, at each time stept, agent not only predicts the next action based on the current overall goal, but also infers the sequence of historical actions that resulted in the present state. 

\noindent \textbf{Thinking format.} 
% 为了在进行推理优化时更准确地将思考过程与预测结果进行分离，以及更加无缝地与function call相集成，我们保留了分隔符对模型输出的不同部分进行分隔。形式上来说，模型的输出格式如下：
To more precisely distinguish the reasoning process from action prediction during inference optimization, and to facilitate seamless integration with function calls, we employ explicit separators to demarcate different sections of the model output. Specifically, the model’s output is structured in the following format:
\begin{tcolorbox}[colframe=black, colback=gray!20, coltitle=black, fonttitle=\bfseries] 
\texttt{<observation>} Observation \texttt{</observation>} \\
\texttt{<think>} Thought \texttt{</think>} \\
\texttt{<tool\_call>} Actions \texttt{</tool\_call>} \\
\texttt{<conclusion>} Conclusion \texttt{</conclusion>}
\end{tcolorbox}

\subsubsection{Distillation Data}
% 人工标注真实场景下的长轨迹数据耗费大量成本，因此，我们构建了一个全自动化的轨迹收集流程，包括三个阶段：（1）真实任务引导的自动化任务生成（2）仿真环境下的自动化任务执行（3）基于多模态大模型投票的轨迹结果判断。经过数据清洗和拆分，我们最终获得500k单步轨迹数据用于训练
Manual labeling of long trajectory data in real scenarios is costly. Therefore, we construct a fully automated trajectory collection process, which includes three stages: (1) Automated task generation guided by real tasks, (2) Automated task execution in simulation environment, and (3) Trajectory result judgment based on VLM voting. After data cleaning and splitting, we finally obtain 500k single-step trajectory data for training.

\noindent \textbf{Task Generation.} 
% 由于缺乏对于具体场景的感知，直接用VLM生成任务可能会导致不明确、不可执行任务的产生。为了解决这个问题，我们将osworld中的369个已有任务的初始状态作为提示，使用GPT-4o扩展生成了更多相关但不同的任务。为了避免初始状态不同导致的任务偏移，我们还为每个任务配备了相应的初始状态。
Directly generating tasks with a VLM, without contextual awareness of specific scenarios, often results in unclear or unexecutable tasks. To mitigate this issue, we utilized the initial states of 369 existing tasks in Osworld as prompts to generate additional, related yet distinct tasks using the GPT-4o extension. Furthermore, to prevent task misalignment caused by varying initial states, each generated task is paired with its corresponding initial state.

\noindent \textbf{Trajectory Distillation.} 
Building on the multi-domain tasks generated by the VLM, we integrated state-of-the-art GUI agent models into the Osworld simulation environment and implemented a concurrent trajectory distillation pipeline. For each task, the model is allowed up to $n$ attempts, with the reasoning process and specific actions of each step recorded. The complete execution trajectory and task details are then evaluated by a VLM to determine whether the task was successfully completed. Additionally, we tracked the number of attempts for each task: data that succeeded in a single attempt were utilized for supervised fine-tuning (SFT), while data requiring multiple attempts were identified as challenging cases and used for GRPO training.
 
\noindent \textbf{Voting Samples.} 
% When the model performs multiple interactions in the virtual environment to generate a complete exploration trajectory, a trajectory judgment model is needed to score whether the trajectory is successfully executed. As shown in Figure 2, the main design principles of the trajectory result discriminator are as follows: (1) Visual as the center, gui agent is different from other agents, task execution mainly depends on the interaction of gui interface, so the change of gui interface can be used as the main basis of task execution state. (2) Using a voting mechanism, whether it is SFT or grpo training, a slight change in the quality of the training data will cause the prediction accuracy of the model to fluctuate. Therefore, we adopt a more strict voting consensus mechanism to sample the same trajectory for multiple evaluations, and only assign a positive label to the trajectory when all evaluations are correct. (3) Difficulty classification: Based on the multiple sampling strategy, each task is inferred by the model multiple times, and the difficulty of the sample is graded according to the statistics of the successful times of multiple inference, which is applied to different training stages.
% When the model engages in multiple interactions within the virtual environment to generate a complete exploration trajectory, it is essential to employ a trajectory evaluation model to determine whether the trajectory has been successfully executed.  Figure  ~\cite{fig:fig2}
As illustrated in Figure 2, the trajectory result discriminator is designed according to the following key principles: (1) Visual-Centric Evaluation: GUI agents primarily rely on interactions with the graphical interface. Therefore, changes in the GUI interface serve as the primary indicators of task execution status. (2) Voting Mechanism: Both in supervised fine-tuning (SFT) and GRPO training, even minor variations in training data quality can lead to fluctuations in model prediction accuracy. To ensure robustness, we adopt a stringent voting consensus mechanism, wherein each trajectory is sampled and evaluated multiple times. A trajectory is assigned a positive label only if all evaluations unanimously indicate success. (3) Difficulty Classification: Leveraging the multi-sampling strategy, each task is inferred multiple times by the model. The sample difficulty is then graded based on the number of successful executions across these inferences, enabling targeted application of samples to different training stages.

\begin{figure}[t]
    \centering
    \includegraphics[width=0.85\linewidth]{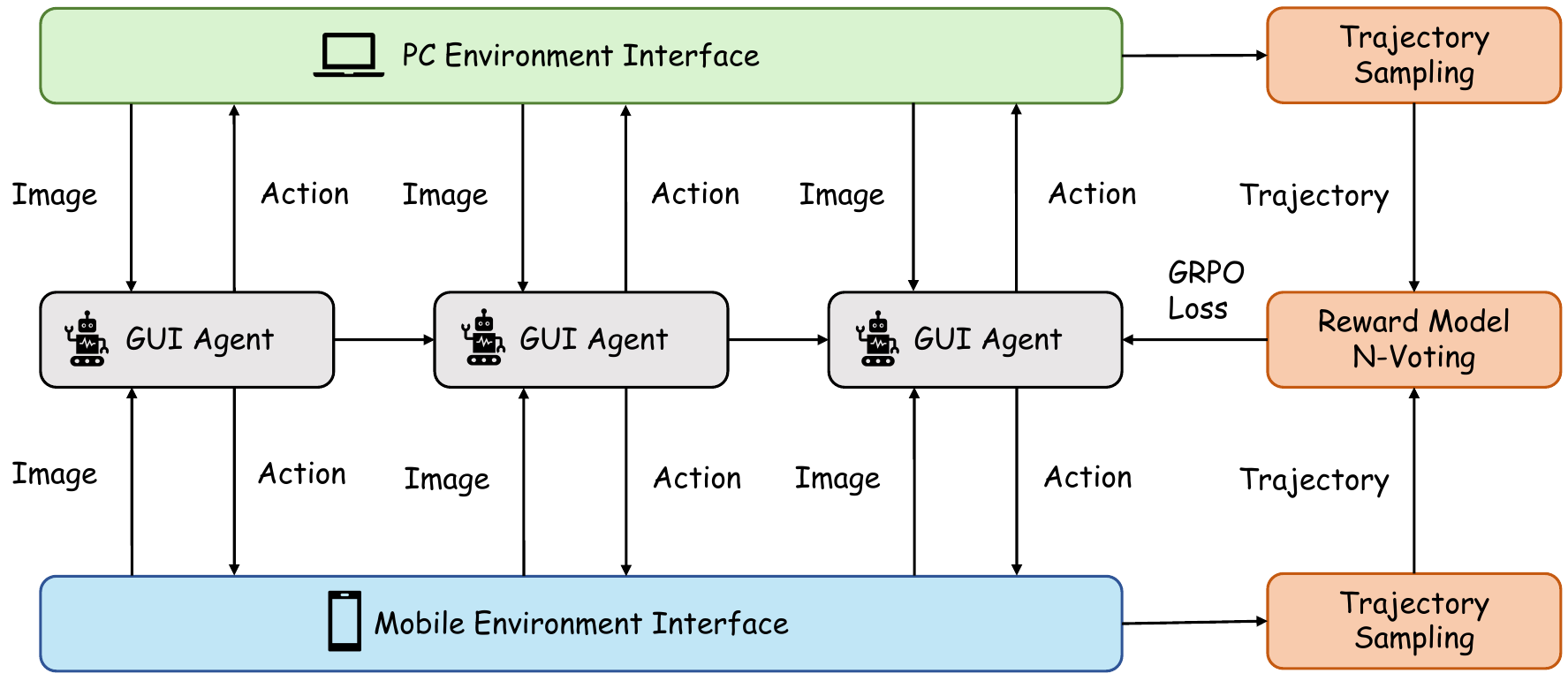}
    \caption{Overall introduction of interactive infrastructure.}
    \label{fig:fig5}
    \vspace{-5pt}
\end{figure}

\subsubsection{General Multimodal Data} 

The general multimodal data serves as a rich repository of fundamental and universal knowledge, intimately interwoven with GUI-related datasets. Recognizing this intrinsic connection, we augment our training regime with image-text pair data sourced from diverse task scenarios such as Optical Character Recognition (OCR), Visual Question Answering (VQA), and Image Captioning. This incorporation aims to bolster the GUI agents' capability to seamlessly comprehend visual content and accurately interpret directive objectives across varied contexts, ultimately fostering a more holistic understanding of task execution dynamics. By leveraging this diverse array of multimodal inputs, we strive to enrich the GUI agents' adaptability and cognitive depth, equipping them to meet increasingly complex interaction demands.
% To overcome these limitations, we augment our training data by integrating a wide range of non-trajectory GUI-related information. This includes element annotations, component attributes, and textual descriptions that provide rich semantic and contextual cues about the interface. Additionally, we incorporate general multimodal data, such as image-text pairs, to further enhance the agent’s ability to perceive, interpret, and reason about diverse GUI scenarios. By systematically fusing these complementary data sources, UItron achieves a deeper and more holistic understanding of user interfaces, significantly improving its perception and reasoning capabilities.

\subsubsection{Manual Annotation.} 
A comprehensive and representative training dataset is essential for developing a robust GUI agent. However, most existing datasets are predominantly focused on English-language applications, leaving a significant gap in coverage for Chinese apps and interfaces. This imbalance limits the agent’s ability to generalize and perform effectively in Chinese application scenarios.
To address this critical shortcoming, we assembled a dedicated team to manually collect operation trajectories specifically targeting top-tier Chinese mobile applications. Our annotation efforts focused on capturing diverse tasks, complex user interactions, and a wide variety of interface designs unique to the Chinese app ecosystem. Through this targeted data collection, we substantially broadened the scope and diversity of our training data, ensuring that UItron is equipped to excel in both English and Chinese application environments.

\subsection{Interactive Infrastructure}
% A scalable, unified interactive infrastructure is essential for reliable data acquisition and effective training of GUI agents. To this end, UItron integrates automation and reinforcement learning into a cohesive system that supports both mobile and PC platforms. This unified infrastructure not only streamlines the entire development process, but also empowers UItron to achieve robust generalization and high performance across diverse application environments.
To facilitate trajectory data collection, online evaluation and RL training, we build an interactive environment connecting both Mobile and PC devices, as shown in Figure 3. 
Specifically, its significance comes from the following three aspects. First, the Mobile and PC interactive environment provides an automated function for recording screenshots and coordinates, which significantly simplifies the difficulty of manually annotating trajectory data and thus accelerates our efficiency in collecting trajectories for Chinese scenarios. Then, the Mobile and PC interactive environment provides a more realistic evaluation environment, simulating the real interaction process between GUI agents and humans. Finally, the Mobile and PC interactive environment provides the execution results of each action output that facilitate online reinforcement learning for the entire trajectory.

\noindent \textbf{Mobile Infra.} 
We build an Android-based cloud real-device environment, which connects multiple genuine Android devices via a server, allowing users to remotely control these smartphones through a web browser. The system is composed of three key components:
\begin{itemize}
    \item Scrcpy: Responsible for streaming the smartphone’s screen content to the browser in real time, similar to live streaming.
    \item Phone-server: Converts user interactions like clicks and swipes made in the browser into touch commands that the smartphone can understand.
    \item Device-agent: Serves as the device management center, integrating the functionalities of the previous two components and providing HTTP interfaces for application installation and device information retrieval.
\end{itemize}

\begin{figure}[t]
    \centering
    \includegraphics[width=0.9\linewidth]{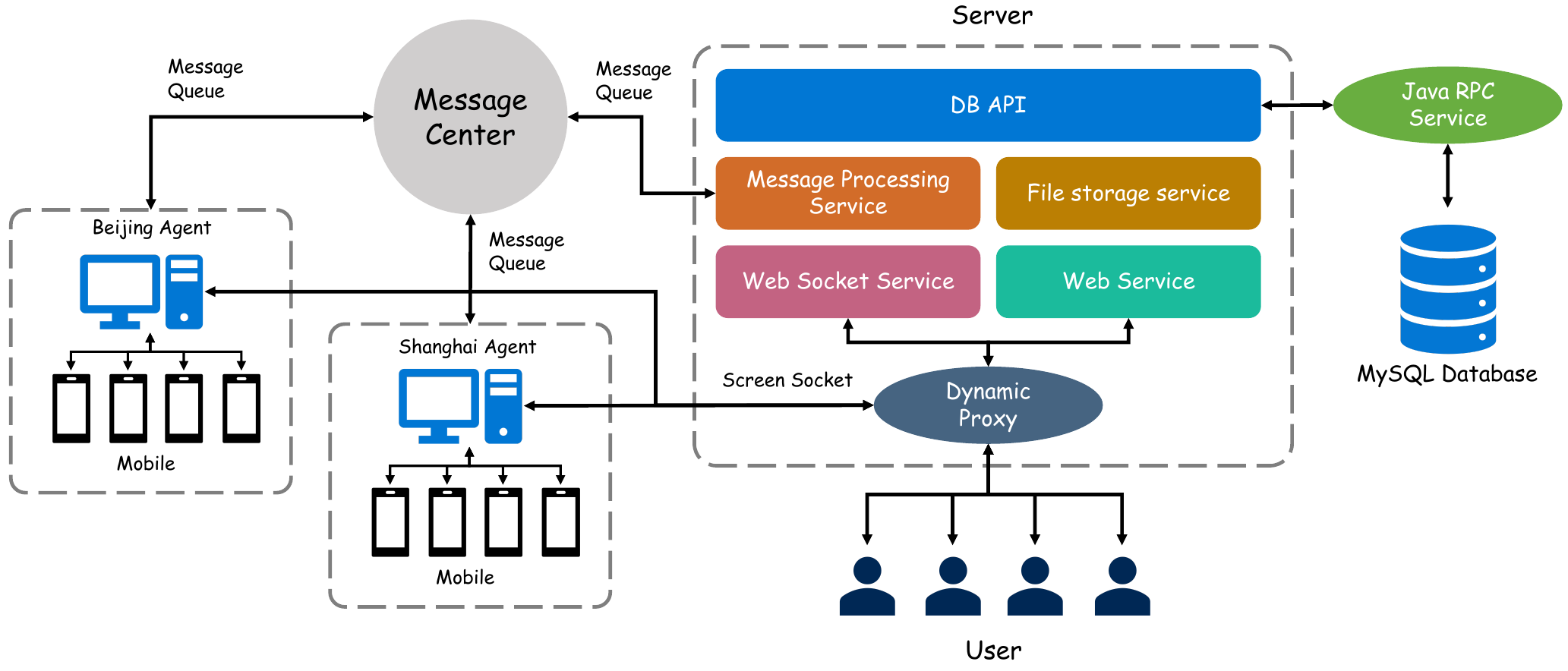}
    \caption{The overall architecture of Mobile infra.}
    \label{fig:arch}
    \vspace{-5pt}
\end{figure}

The architecture follows an Agent/Server model, with the server side handling the user interface and device scheduling, while the Agent side manages the specific smartphone devices. Real-time communication is facilitated via WebSocket, and MySQL is used to store device and user data.
This solution addresses prevalent issues in mobile application testing, such as insufficient device availability, incomplete model coverage, and the need for remote operations.

% Reliable and large-scale trajectory data are fundamental for training effective GUI agents. However, manual data collection is time-consuming, error-prone, and difficult to scale across diverse platforms and application scenarios, ultimately hindering the development of robust models capable of generalizing to real-world environments. To overcome these challenges, we developed an automated data collection tool within our unified interactive infrastructure that systematically records each operation’s screen capture, coordinates, and action type. This infrastructure greatly streamlines the trajectory data acquisition process, ensuring consistency, accuracy, and scalability. By automating data collection across both mobile and PC environments, we efficiently build comprehensive datasets that underpin high-quality model training and evaluation.

\noindent \textbf{PC Infra.} 
We utilize the open-source OSWorld environment ~\cite{OSWorld}, a scalable real computer setting specifically designed for developing multimodal agents capable of executing a wide array of real computer tasks beyond isolated interfaces and applications. This executable environment allows unrestricted keyboard and mouse control over real computer applications, supporting initial task state configuration, execution-based assessment, and interactive learning across major operating systems like Ubuntu, Windows, and macOS. Moreover, it provides the capability to evaluate open-ended computer tasks, encompassing activities from image viewing and software feature integration to programming. Hence, OSWorld serves as a unified real computing environment where users can define their agent tasks without the need to construct simulation environments tailored to specific applications or domains.

% \noindent \textbf{Online Evaluation Environment.} 
% Comprehensive and realistic evaluation is essential for measuring the true performance and generalization ability of GUI agents. However, relying solely on either simulators or real devices can lead to incomplete assessments—simulators may not fully capture real-world complexity, while exclusive use of real devices can be resource-intensive and limit scalability.
% To overcome these limitations, we have constructed a hybrid evaluation environment that seamlessly integrates both simulators and real devices. This platform supports offline trajectory replay for efficient large-scale testing, as well as online agent interaction to capture dynamic, real-world behaviors. By leveraging the strengths of both simulation and real-device evaluation, our environment enables thorough experimentation and robust performance assessment across a wide range of application scenarios.

\subsection{Training Paradigm}
During training, we employ a three-stage training strategy (as shown in Figure 4), in which consists of two SFT stages for perception and planning tasks, as well as a RL stage with curriculum reinforcement learning framework.
In the first stage, the perception task focuses on improving the basic understanding ability of the vision-language model in GUI scenarios, such as grounding, captioning, VQA, and OCR. In the second stage, the planning task concentrates on predicting the next action based on historical actions.
In the final RL stage, the curriculum reinforcement learning framework aims to improve reasoning and exploration capacity via group relative policy optimization algorithm on trajectory data.

\begin{figure}[t]
    \centering
    \includegraphics[width=1.0\linewidth]{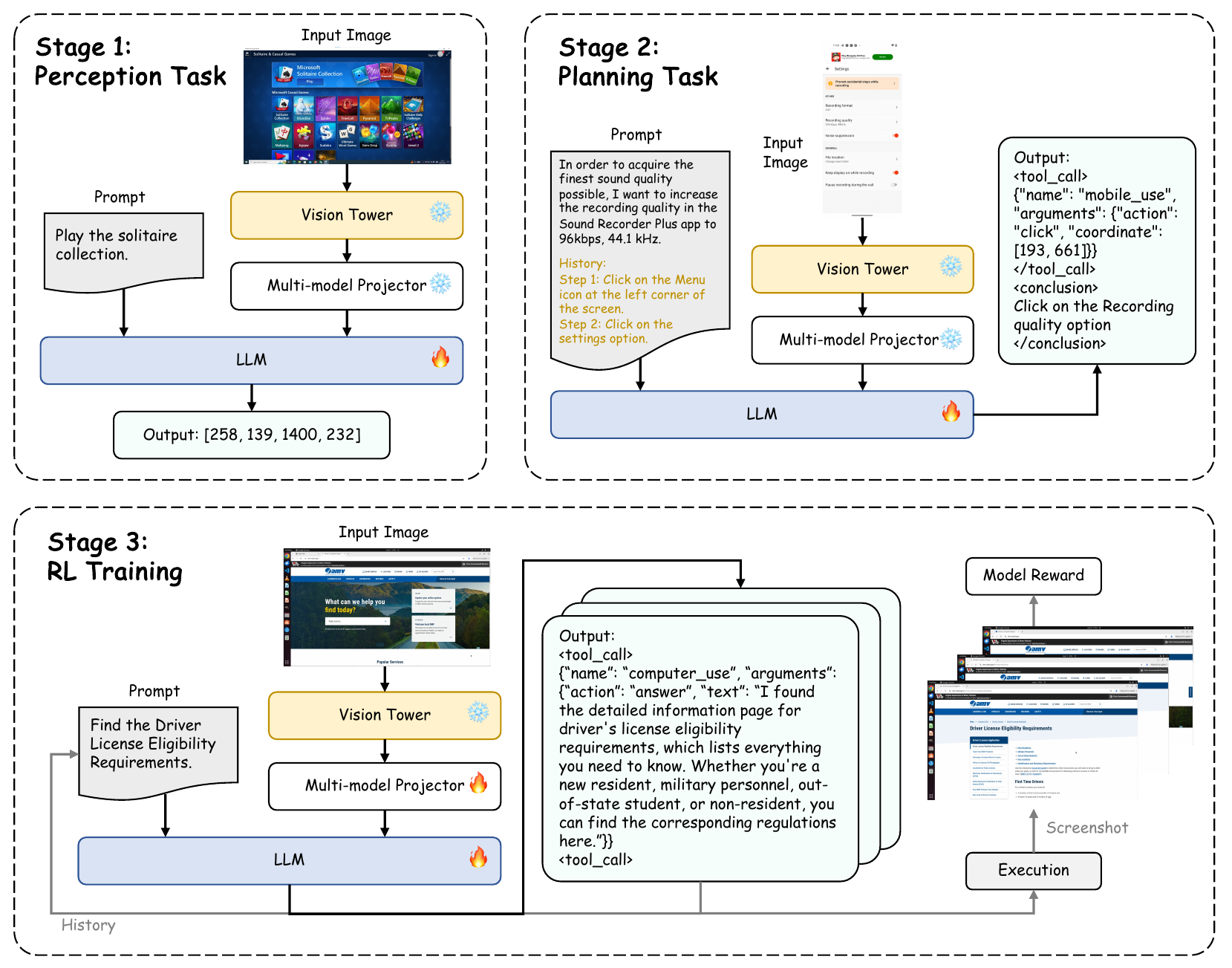}
    \vspace{-15pt}
    \caption{Overall introduction of training paradigm.}
    \label{fig:fig4}
    \vspace{-5pt}
\end{figure}

\subsubsection{Stage 1: Perception Task} 
The perceptual abilities of a GUI agent are fundamental for enabling deep understanding and effective interaction with digital interfaces. Modern digital environments are increasingly complex, with user interfaces containing rich visual elements, diverse layouts, and embedded semantic information. Without robust perception capabilities, an agent would struggle to interpret the structure, content, and intent behind various UI components, thereby limiting its effectiveness in real-world applications.

To address this critical challenge, we initially enhance UItron's perceptual ability in the first stage by fine-tuning it on a wide range of GUI perception scenarios. The goal of this fine-tuning is to systematically strengthen UItron’s ability to recognize and interpret interface elements, ensuring a deeper and more precise understanding of digital UIs. In particular, we focus on four core perception tasks: grounding, captioning, VQA, and OCR. Grounding enables the agent to accurately localize and associate semantic labels with interface components, establishing a clear mapping between visual regions and their meanings. Captioning facilitates the generation of natural language descriptions for UI layouts and elements, allowing the agent to summarize and communicate interface structures effectively. Furthermore, VQA empowers the agent to answer queries about the interface by integrating both visual and semantic cues, supporting interactive and context-aware understanding. OCR extracts embedded textual information from the interface, ensuring that no detail is overlooked and that all relevant data is accessible for downstream reasoning.

Through mastering these perception tasks, the fine-tuned UItron attains a holistic and nuanced understanding of user interfaces. This comprehensive perceptual foundation not only enhances its ability to interpret complex digital environments, but also lays a solid groundwork for advanced reasoning, planning, and autonomous interaction in subsequent stages. 

\subsubsection{Stage 2: Planning Task} 

In this stage, training with planning tasks aims to output the predicted actions defined in Equation (1), then optimize the output using a generative loss in an auto-regressive manner.
Effective planning is a key capability that enables a GUI agent to execute purposeful actions and navigate complex digital environments. 

The centra idea of planning task is to capture next actions for forward planning and historical actions for backtracking.
To this end, we construct two types of training data in the planning task, one for forward planning and the other for backward backtracking. 
The forward planning follows the message organization approach defined in Equation (1), which inputs historical actions $(a_1, a_2,...,a_{n-1})$ and the environment observation $o_n$ to output next action $a_n$.
In contrast, the backward backtracking follows the message organization approach defined
as follows:
\begin{equation}\label{eq_Future}
    a_{n-1},a_n = \mathbf{M}_\theta(T, (a_1, a_2,...,a_{n-2}),o_{n-1},o_n),
\end{equation}
Here the main difference is that the previous action $a_{n-1}$ is not provided in the input, while the agent need to predict $a_{n-1}$.

% To endow UItron with advanced planning performance, we introduce a second fine-tuning stage that centers on two complementary patterns. 
% The first pattern constructs action plans based solely on the current UI snapshot, enabling UItron to comprehend instruction tasks and the present UI state, reason about task requirements, and devise the next steps toward task completion. In contrast, the second paradigm integrates historical actions with present observations, leveraging temporal context to inform future decisions. This approach empowers the agent to recognize patterns, reflect on prior interactions, and adapt its planning in response to changes or feedback. Importantly, both training paradigms are applied across mobile and desktop operating system scenarios, exposing UItron to a diverse array of interface layouts, interaction modalities, and platform-specific constraints. Through learning to navigate and strategize within these varied environments, UItron develops the versatility required to tackle a wide spectrum of real-world applications.

% With these planning skills in place, UItron transitions from passive observation to active problem-solving. It becomes proficient not only at describing what it sees, but also at reasoning about what to do next, weighing alternatives, and executing multi-step strategies. Ultimately, such capabilities set the stage for reinforcement learning, where the agent can further hone its decision-making, optimize its actions, and autonomously pursue increasingly sophisticated objectives.

\subsubsection{Stage 3: Curriculum Reinforcement Learning}
To enhance the reasoning ability, UItron develops a curriculum reinforcement learning framework for performing group relative policy optimization (GRPO) \cite{shao2024deepseekmath}algorithm on trajectory data.
It first computes dense rewards from the action steps in the offline environment (simple), and then computes the task-level reward for the trajectory data in the online environment (complex).

\noindent \textbf{GRPO}

We adapt the Group Relative Policy Optimization (GRPO) \cite{shao2024deepseekmath} algorithm for RL.
For each input $(x, y)$, the policy $\pi_\theta$ samples a group of $G$ candidate responses $\{o_i\}_{i=1}^G$.
\begin{align}
\mathcal{J}_{GRPO}(\theta) = & \mathbb{E}_{\substack{(x,y) \sim \mathcal{D}, \{o_i\}_{i=1}^G \sim \pi_{\theta_{\text{old}}} (O \mid x)}} \biggl[ \frac{1}{G} \sum_{i=1}^G \min \biggl(\frac{\pi_\theta(o_i \mid x)}{\pi_{\theta_{\text{old}}}(o_i \mid x)} A_i,\\
& \operatorname{clip}\left(\frac{\pi_\theta(o_i \mid x)}{\pi_{\theta_{\text{old}}}(o_i \mid x)}, 1-\varepsilon, 1+\varepsilon\right) A_i\biggr) \notag  - \beta \mathbb{D}_{KL}(\pi_\theta \| \pi_{\mathrm{SFT}}) \biggr]
\end{align}
where $\varepsilon$ and $\beta$ are hyperparameters, and $\pi_{\mathrm{SFT}}$, $\pi_{\theta}$, and $\pi_{\theta_{\text{old}}}$ are the model after SFT, the optimized model and the old policy model.
The group-normalized advantage for the $i$-th response is:
\begin{equation}
A_i=\frac{r_i-\operatorname{mean}\left(\left\{r_1, r_2, \cdots, r_G\right\}\right)}{\operatorname{std}\left(\left\{r_1, r_2, \cdots, r_G\right\}\right)}
\end{equation}

\noindent \textbf{Offline RL} 
Since successful trajectories of GUI agents in online environments are usually rare, the offline RL collects rewards for each action step in the offline environment to avoid sparse rewards. 
For each action prediction, it generates multiple candidate actions for the same input and calculates the GRPO loss to improve reasoning and exploration capabilities.

\noindent \textbf{Online RL} 
The online RL is built on the interactive infrastructure shown in Figure 3. It collects rewards for the entire trajectory through the rollout dialogues in an online environment.
For each task, online RL allows the model to freely explore all possible action plans until it reaches the maximum number of steps or generates an end signal.
To this end, online RL utilizes the advanced vision-language model as a scoring model to evaluate whether the task is completed based on the entire trajectory and task, and outputs a reward signal of 0 or 1.
To improve the credibility of rewards in the RL process, we incorporate multiple scoring models from different vision-language models. Besides, we also strictly filter the trajectories that are predicted correctly by multiple scoring models simultaneously.
Finally, the online RL calculates the trajectory-level GRPO loss based on multiple sampled trajectories, thereby improving the exploration ability in the online environment.

% \subsection{Adaptation to Chinese Application Scenarios}
% % \paragraph{Large-scale Chinese Trajectory Collection} To address the lack of Chinese capabilities in existing solutions, UItron specifically collects over one million operation trajectories covering diverse scenarios in top Chinese mobile apps.

% \paragraph{Dedicated Chinese Task Evaluation} Offline and online evaluation platforms are established to assess the model’s perception, localization, and operation proficiency in Chinese application environments, driving real-world adoption in the Chinese ecosystem.

\noindent \textbf{Summary}
Finally, we produce two versions named UItron and UItron-RL, both of which are based on the Qwen25-VL model structure, but with different parameter weights.
The former is obtained after training in stages 1 and 2, while the latter is obtained after reinforcement training in stage 3. In the experiments, we report the results of uitron-RL in all online environments, and report the results of UItron in other offline scenarios.

\begin{table*}[]
\centering
\begin{adjustbox}{max width=0.9\textwidth}
\begin{tabular}{lllcc}
\toprule
\textbf{Benchmarks} & \textbf{Task} & \textbf{Platform} & \textbf{Metric} & \textbf{\# Test Samples} \\
\midrule
\multirow{5}{*}{VisualWebBench}& Element Grounding & Web & Prediction Accuracy & 413 \\
& Action Grounding & Web & Prediction Accuracy & 103 \\
& Element OCR & Web & ROUGE-L & 245 \\
& Heading OCR & Web & ROUGE-L & 46 \\
& Web QA & Web & SQuAD-F1 & 314 \\
\midrule
RefExp & Task Grounding & Web & Accuracy ($IoU \ge 0.5$) & 1000 \\
WidgetCap & Element Caption & Mobile & CIDEr & 1000 \\
WebSRC & WebQA & Web & SQuAD-F1 & 1000 \\
\bottomrule
\end{tabular}
\end{adjustbox}
\caption{Details of GUI perception benchmarks. All evaluation data is structured as single-round conversations during experimental.}
\label{tab:perception_metric}
\end{table*}

\begin{table*}[]
\centering
\begin{adjustbox}{max width=0.9\textwidth}
\begin{tabular}{llccc}
\toprule
\textbf{Benchmarks}         & \textbf{Platform} & \textbf{\#Test Episodes} & \textbf{\#Test Samples} & \textbf{History} \\ \midrule
ScreenSpot & Mobile\&Desktop\&Web & - & 1272 & \\
ScreenSpot-V2 & Mobile\&Desktop\&Web & - & 1272 & \\
\midrule
AndroidContorl-Low &Mobile &1,000 &5,477 &          \\
AndroidContorl-High &Mobile &1,000 &5,477 & \checkmark         \\
\midrule
GUI-Odyssey-Random &Mobile &1,933 &29,426 & \checkmark       \\
GUI-Odyssey-App &Mobile &1,139 &17,455 & \checkmark      \\
GUI-Odyssey-Device &Mobile &1,262 &18,967 & \checkmark        \\ 
GUI-Odyssey-Task &Mobile &1,016 &17,920 & \checkmark     \\ 
\midrule
OSWorld &Desktop\&Web &369 &- & \checkmark     \\ 
AndroidWorld &Mobile &116 &- & \checkmark  \\
MobileMiniWob &Web &92 &- & \checkmark  \\
\bottomrule
\end{tabular}
\end{adjustbox}
\caption{Details of the agentic benchmarks. ``Test episodes'' refers to the number of trajectory data used for evaluation, while ``Test samples'' represents the total number of individual step data contained within all trajectories. ``History'' indicates whether the historical information of previous actions is provided in the model input.}
\label{tab:benchmarks}
\end{table*}

\section{Experiments}

We carry out extensive experiments covering scenarios including GUI perception, grounding, offline planning, and online planning. In particular, we also built our own Chinese scenario evaluation and conduct experiments to explore the improvement of Chinese capabilities.

\subsection{ Evaluation of GUI Perception}
\paragraph{VisualWebBench.}
We evaluate our model's screen perception capabilities on VisualWebBench~\cite{liu2024visualwebbench}, a comprehensive benchmark containing multiple website-based tasks. For the Grounding Tasks, we measure prediction accuracy by requiring the agent to select correct answers from set of masks (SoM) on screenshots. 

\paragraph{Complicated Perceptual Benchmarks.}
To evaluate the model's ability to comprehend abstract instructions, we follow \cite{liu2024harnessing} by assessing visual grounding of natural language-described elements through RefExp~\cite{bai2021uibert} and testing the reverse task of element captioning on WidgetCap~\cite{li2020widget}. We additionally evaluate on general Web QA task of WebSRC~\cite{chen2021websrc}, requiring textual and structural understanding of GUI elements, for further assessing comprehensive perceptual capabilities. 

\begin{table*}[]
\centering
\begin{adjustbox}{max width=0.8\textwidth}
\begin{tabular}{lccccc}
\toprule
\textbf{Method} & \textbf{\makecell{Element\\Grounding}} & \textbf{\makecell{Action\\Grounding}} & \textbf{\makecell{Element\\OCR}} & \textbf{\makecell{Heading\\OCR}} & \textbf{Web QA} \\
\midrule
GPT-4o & 79.91 & 86.41 & 79.42 & 64.57 & 77.44 \\
Qwen2.5-VL & 82.81 & 77.67 & 95.72 & 66.82 & 80.23 \\
MutiUI & 75.92 & 36.66 & - & - & - \\
% UI-TARS-7B & - & - & - & - & 79.7 \\
UI-TARS & 96.13 & 92.23 & 92.53 & 70.63 & 79.7 \\
% UI-TARS-72B & - & - & - & - & \textbf{82.8} \\
\midrule
UItron-7B & 94.67 & 94.07 & 95.36 & 65.50 & 77.70 \\
UItron-72B & 96.37 & 94.17 & 95.56 & 72.15 & 80.49 \\
\bottomrule
\end{tabular}
\end{adjustbox}
\caption{Comparative results on VisualWebBench~\cite{liu2024visualwebbench}.}
\label{tab:perception1}
\end{table*}

\begin{table*}[]
\centering
\begin{adjustbox}{max width=0.75\textwidth}
\begin{tabular}{lccccc}
\toprule
\multirow{2}{*}{\textbf{Method}} & \textbf{Task Grounding} & \textbf{Element Caption} & \textbf{Web QA} \\
&\textbf{(RefExp)} & \textbf{(WidgetCap)} &\textbf{(WebSRC)} \\
\midrule
GPT-4o & --& 61.02 & 78.60  \\
Qwen2.5-VL & 6.55 & 58.45 & 91.2 \\
MutiUI & 43.56 & 72.73 & 82.9 \\
% UI-TARS-7B & - & - & \textbf{93.6} \\
UI-TARS & -- & 82.48 & 93.6 \\
% UI-TARS-72B & - & - & 92.1 \\
\midrule
UItron-7B & 51.40 & 77.55 & 89.7 \\
UItron-72B & 59.20 & 87.44& 93.24 \\
\bottomrule
\end{tabular}
\end{adjustbox}
\caption{Comparative results on RefExp~\cite{bai2021uibert},WidgetCap~\cite{li2020widget} and WebSRC~\cite{chen2021websrc}.
Note that GPT-4o and UI-TARS are failed to evaluate due to the invalid output format in task grounding (RefExp). }
\label{tab:perception2}
\end{table*}

\paragraph{Baseline Models.}
We compare our UItron with SOTA models in both understanding and GUI operation task. Among general VLLMs, we use GPT-4o~\cite{hurst2024gpt} and Qwen2.5-VL~\cite{bai2025qwen2} as our baseline for their powerful understanding capabilities in general task understanding; among GUI-related VLLMs, we compare our UItron with MultiUI~\cite{liu2024harnessing} and UI-TARS~\cite{qin2025ui}, the former is specialized in GUI understanding while the later one is the SOTA model in GUI tasks.

As shown in Tables~\ref{tab:perception1} and \ref{tab:perception2}, our UItron demonstrates superior performance on perceptual tasks, establishing crucial groundwork for subsequent planning and reasoning capabilities essential for GUI task execution. This effectiveness stems from the limited understanding data employed in both training stages 1 and 2, which not only mitigates the spurious forgetting issue~\cite{zheng2025spurious} that degrades baseline VLLM's original comprehension, but also enhances GUI-specific understanding. This results indicate that maintaining the generalist model's capabilities relevant to the the downstream task while developing specialized skills is critical for creating effective specialist agents.
Details of the benchamrks are listed in Table~\ref{tab:perception_metric}.

\begin{table*}[h]
\centering
\begin{adjustbox}{max width=0.95\columnwidth}
\begin{tabular}{@{}p{1.5cm}lccccccp{0.8cm}@{}}
\toprule
\multicolumn{2}{l}{\multirow{2}{*}{\textbf{Method}}} & \multicolumn{2}{c}{\textbf{Mobile}} & \multicolumn{2}{c}{\textbf{Desktop}} & \multicolumn{2}{c}{\textbf{Web}} & \multirow{2}{*}{\textbf{Avg}} \\
\cmidrule{3-8}
& & \textbf{Text} & \textbf{Icon/Widget} & \textbf{Text} & \textbf{Icon/Widget} & \textbf{Text} & \textbf{Icon/Widget} & \\
\midrule
\multicolumn{9}{l}{\textit{Agent Framework}} \\
\midrule
\multirow{2}{*}{GPT-4o} & SeeClick &81.0 &59.6 &69.6 &33.6 &43.9 &26.2 &52.3 \\
 & UGround &93.4 &76.9 &92.8 &67.9 &88.7 &68.9 &81.4 \\
\midrule
\multicolumn{9}{l}{\textit{Agent Model}} \\
\midrule
\multicolumn{2}{l}{GPT-4o} &20.2 &24.9 &21.1 &23.6 &12.2 &7.8 &18.3 \\
\multicolumn{2}{l}{Claude} &- &- &- &- &- &- &83.0 \\
\multicolumn{2}{l}{Gemini 2.0} &- &- &- &- &- &- &84.0 \\

\multicolumn{2}{l}{Qwen2.5-VL-7B} &- &- &- &- &- &- &84.7 \\
% \multicolumn{2}{l}{CogAgent} &67.0 &24.0 &74.2 &20.0 &70.4 &28.6 &47.4 \\
% \multicolumn{2}{l}{SeeClick} &78.0 &52.0 &72.2 &30.0 &55.7 &32.5 &53.4 \\
\multicolumn{2}{l}{UGround} &82.8 &60.3 &82.5 &63.6 &80.4 &70.4 &73.3 \\
\multicolumn{2}{l}{Aria-UI} &92.3 &73.8 &93.3 &64.3 &86.5 &76.2 &82.4 \\
\multicolumn{2}{l}{OS-Atlas} &93.0 &72.9 &91.8 &62.9 &90.9 &74.3 &82.5 \\
\multicolumn{2}{l}{AGUVIS-7B} &95.6 &77.7 &93.8 &67.1 &88.3 &75.2 &84.4 \\
\multicolumn{2}{l}{AGUVIS-72B} &94.5 &85.5 &95.4 &77.9 &91.3 &85.9 &89.2 \\
\multicolumn{2}{l}{UI-TARS-7B} &94.5 &85.2 &95.9 &85.7 &90.0 &83.5 &89.5 \\ 
\multicolumn{2}{l}{UI-TARS-72B} &94.9 &82.5 &89.7 &88.6&88.7 &85.0 &88.4 \\ 
\midrule
\multicolumn{2}{l}{UItron-7B} &94.1 &83.8 &94.8 &73.6 &92.2 &81.1 &87.7 \\ 
% \multicolumn{2}{l}{UItron-72B} &94.5 &88.6 &95.9 &85.7 &93.9 &88.3 &91.6\\ 
\multicolumn{2}{l}{UItron-72B} &94.5 &88.2 &96.9 &79.2 &93.0 &85.4 &90.3\\ 
\bottomrule 
\end{tabular}
\end{adjustbox}
\label{tab:grouding}
\caption{Comparison of different baseline methods on ScreenSpot~\cite{cheng2024seeclick}.}
\end{table*}

\begin{table*}[h]
\centering
\begin{adjustbox}{max width=0.95\columnwidth}
\begin{tabular}{@{}p{1.5cm}lccccccp{0.8cm}@{}}
\toprule
\multicolumn{2}{l}{\multirow{2}{*}{\textbf{Method}}} & \multicolumn{2}{c}{\textbf{Mobile}} & \multicolumn{2}{c}{\textbf{Desktop}} & \multicolumn{2}{c}{\textbf{Web}} & \multirow{2}{*}{\textbf{Avg}} \\
\cmidrule{3-8}
& & \textbf{Text} & \textbf{Icon/Widget} & \textbf{Text} & \textbf{Icon/Widget} & \textbf{Text} & \textbf{Icon/Widget} & \\
\midrule
\multicolumn{9}{l}{\textit{Agent Framework}} \\
\midrule
\multirow{2}{*}{GPT-4o} & SeeClick &85.2 &58.8 &79.9 &37.1 &72.7 &30.1 &63.6 \\
 & OS-Atlas-4B &95.5&75.8 &79.4 &49.3 &90.2 &66.5 &79.1 \\
  & OS-Atlas-7B &96.2 &83.4 &89.7 &69.3 &94.0 &79.8 &87.1 \\
\midrule
\multicolumn{9}{l}{\textit{Agent Model}} \\
\midrule

\multicolumn{2}{l}{SeeClick} &78.4 &50.7 &70.1 &29.3 &55.2 &32.5 &55.1 \\
\multicolumn{2}{l}{OS-Atlas-4B} &87.2 &59.7 &72.7 &46.4 &85.9 &63.1 &71.9 \\
\multicolumn{2}{l}{OS-Atlas-7B} &95.2 &75.8 &90.7 &63.6 &90.6 &77.3 &84.1 \\
\multicolumn{2}{l}{UI-TARS-7B} &96.9 &89.1 &95.4 &85.0 &93.6 &85.2 &91.6 \\ 
\multicolumn{2}{l}{UI-TARS-72B} &94.8 &86.3 &91.2 &87.9 &91.5 &87.7 &90.3 \\ 
\midrule
\multicolumn{2}{l}{UItron-7B} &96.9 &88.2 &92.8 &71.4 &91.5 &80.8 &88.4 \\ 
\multicolumn{2}{l}{UItron-72B} &95.5 &90.5  &99.0 &80.0 &94.0 &87.7 &92.0\\ 
\bottomrule 
\end{tabular}
\end{adjustbox}
\label{tab:grouding}
\caption{Comparison of different baseline methods on ScreenSpot-V2~\cite{cheng2024seeclick}.}
\end{table*}

\subsection{Evaluation of GUI Grounding}
\paragraph{ScreenSpot.} We use ScreenSpot~\cite{cheng2024seeclick} to assess the fundamental GUI-understanding and element-grounding accuracy of GUI-agent models. The ScreenSpot benchmark comprises more than 600 screenshots and 1,200 instructions, spanning multiple platforms—iOS, Android, macOS, Windows, and web pages. We report separate results for Text and Icon/Widget elements on the Mobile, Desktop, and Web splits of ScreenSpot, together with the micro accuracy aggregated across all platforms.

\paragraph{ScreenSpot-V2.} 
Similar with ScreenSpot~\cite{cheng2024seeclick}, we also employ ScreenSpot-V2~\cite{wuatlas} for evaluation, which is a GUI benchmark that advances from basic recognition to cross-modal reasoning. This enhanced version better reflects real-world complexity through optimized annotations, expanded task types, and improved data diversity. The benchmark contains 1,272 instructional samples paired with 756 images, drawing from data sources similar to ScreenSpot.

The experimental results in Table~\ref{tab:grouding} demonstrate that UItron exhibits impressive leading GUI grounding performance across all platforms. This advantage is primarily attributed to UItron's adoption of data engineering specifically tailored for GUI agents, which provides high-quality and well-defined datasets for model training. Furthermore, the parameter scaling experiments of UItron indicate that, with sufficient and high-confidence training data, the model's grounding capability is further enhanced as its scale increases. Compared with state-of-the-art model (\textit{i.e.}, UI-TARS) that additionally utilize internal data, UItron-72B relies solely on open-source data, achieves a 2.1\% improvement in micro grounding accuracy.

\begin{table*}[t]
\centering
\begin{adjustbox}{max width=0.8\width}
\begin{tabular}{@{}p{1.5cm}lccccccccccc@{}}
\toprule
\multicolumn{2}{l}{\multirow{2}{*}{\textbf{Method}}}
& \multicolumn{3}{c}{\textbf{AndroidControl-Low}} 
& \multicolumn{3}{c}{\textbf{AndroidControl-High}} 
& \multicolumn{3}{c}{\textbf{GUI Odyssey}} & \multirow{2}{*}{\textbf{Avg}} \\ 
\cmidrule(l){3-5} \cmidrule(l){6-8} \cmidrule(l){9-11}
& & \textbf{Type} & \textbf{Grounding} & \textbf{SR} 
  & \textbf{Type} & \textbf{Grounding} & \textbf{SR} 
  & \textbf{Type} & \textbf{Grounding} & \textbf{SR}\\ 
\midrule
\multicolumn{11}{l}{\textit{Agent Framework}} \\ \midrule
\multirow{2}{*}{GPT-4o} & SeeClick & - & - & 52.8 & - & - & 41.8 & - & - & -  & - \\
& UGround & - & - & 62.4 & - & - & 48.4 & - & - & - & - \\ \midrule
\multicolumn{11}{l}{\textit{Agent Model}} \\ \midrule
\multicolumn{2}{l}{Claude} &74.3 &0.0 &19.4 &63.7 &0.0 &12.5 &60.9 &0.0 &3.1 & 26.0\\
\multicolumn{2}{l}{GPT-4o} &74.3 &0.0 &19.4 &66.3 &0.0 &20.8 &34.3 &0.0 &3.3 & 24.3\\
\multicolumn{2}{l}{InternVL-2} &90.9 &84.1 &80.1 &84.1 &72.7 &66.7 &82.1 &55.5 &51.5 & 74.2 \\
% Qwen-2VL-7B &91.9 &86.5 &82.6 &83.8 &77.7 &69.7 &83.5 &65.9 &60.2 \\
\multicolumn{2}{l}{SeeClick} &93.0 &73.4 &75.0 &82.9 &62.9 &59.1 &71.0 &52.4 &53.9 & 69.3\\
\multicolumn{2}{l}{Aria-UI} &- &87.7 &67.3 &- &43.2 &10.2 &- & 86.8 &36.5 & -\\
% \multicolumn{2}{l}{OS-Atlas-4B} &91.9 &83.8 &80.6 &84.7 &73.8 &67.5 &83.5 &61.4 &56.4 \\
\multicolumn{2}{l}{OS-Atlas} &93.6 &88.0 &85.2 &85.2 &78.5 &71.2 &84.5 &67.8 &62.0 & 79.6 \\
\multicolumn{2}{l}{AGUVIS} &- &- &80.5 &- &- &61.5 &- &- &- & -\\
\multicolumn{2}{l}{UI-TARS-7B} &98.0 &89.3 &90.8 &83.7 &80.5 &72.5 &94.6 &90.1 &87.0 & 87.4 \\
\multicolumn{2}{l}{UI-TARS-72B} &98.1 &89.9 &91.3 &85.2 &81.5 &74.7 &95.4 &91.4 &88.6 & 88.5 \\
\midrule
\multicolumn{2}{l}{UItron-7B} & 95.8&90.7 &88.2 & 96.1 & 87.1 &86.9 & 95.3 & 85.7&84.8 & 90.1 \\
\multicolumn{2}{l}{UItron-72B} &96.5 &97.1 &92.8 &96.7  & 94.2 &92.4 & 94.4 &86.3 & 86.1 & 92.9\\
\bottomrule
\end{tabular}
\end{adjustbox}
\caption{Comparative results on AndroidControl-Low~\cite{li2024effects}, AndroidControl-High~\cite{li2024effects} and GUI-Odyssey~\cite{lu2024gui}.}
\label{tab:offline}
\end{table*}

\subsection{Evaluation of Offline Planning}
\paragraph{AndroidControl.}
AndroidControl~\cite{li2024effects} is a benchmark for evaluating the planning and action-execution capabilities of GUI agents on Android devices. It contains 15,283 episodes of everyday tasks across 833 distinct applications, making it the most diverse UI-control dataset to date. Following standard practice, we report results under two settings. AndroidControl-Low: At every step the agent receives a screenshot together with a natural-language description of the required action and must predict both the action type and its exact parameters. AndroidControl-High: Only the high-level task goal and the current screenshot are provided at each step. The agent must autonomously plan the entire procedure and output the correct action together with its parameters. Following OS-Atlas~\cite{wuatlas}, we reserve 1,000 episodes as an out-of-domain evaluation set and report the action-type accuracy, grounding accuracy, and average step success rate.

General-purpose LLMs such as GPT-4o and Claude demonstrate reasonable action-type accuracy in the Low setting, but their grounding accuracy is essentially zero and their step success rates are very low, indicating a lack of fine-grained perception and UI localization capability. In contrast, specialized GUI models like SeeClick, Aria-UI, OS-Atlas, AGUVIS, and UI-TARS exhibit a clear advantage, achieving substantially higher scores across all metrics. This demonstrates the superiority of dedicated GUI models in both accurately identifying UI elements and executing precise actions. Notably, our model, UItron, surpasses all others: UItron-72B achieves the highest grounding and step success rates in both Low and High settings, showcasing exceptional performance not only in guided UI action execution but also in autonomous planning. This underscores the critical importance of our model's unified approach to perception, grounding, and planning, enabling robust and generalizable UI control.

\paragraph{GUI-Odyssey.}
GUI-Odyssey~\cite{lu2024gui} is used for evaluating cross-app navigation agents, surpassing the limitation of other benchmarks that are restricted to a single app. It consists of 7,735 episodes, six types of cross-app tasks, 201 apps, and 1.4k app combinations. GUI-Odyssey-Random/Task/Device/App are four different test subsets, with statistics shown in Table~\ref{tab:benchmarks}. It aims to assess the generalization ability of autonomous GUI agents across different applications, tasks, and device setups. Following OS-Atlas~\cite{wuatlas}, we report the macro average performance across these subsets.

The challenge of cross-app navigation exposes even greater limitations in general-purpose LLMs, with GPT-4o and Claude displaying poor performance in both grounding and step success rate, and action-type accuracy dropping further compared to single-app scenarios. Specialized GUI models again demonstrate their superiority, with SeeClick, Aria-UI, and OS-Atlas showing solid results, and UI-TARS achieving state-of-the-art performance. Notably, UItron remains highly competitive, achieving results close to those of UI-TARS in most metrics. While UItron may perform slightly below UI-TARS on certain cross-app tasks, it consistently demonstrates top-tier results when considering both AndroidControl and GUI-Odyssey benchmarks together, highlighting its overall superiority in comprehensive UI understanding and control. UItron’s strong performance across diverse tasks and app combinations underscores its robust generalization ability and reliability, making it one of the most effective agents for complex, real-world UI navigation tasks.

\subsection{Evaluation of Online Planning}

\paragraph{OSWorld.}
We use OSWorld~\cite{OSWorld} to evaluate the performance of GUI agent models as online agents on personal computer (PC) platforms. OSWorld is a real computer environment that supports multimodal agents in task setup and execution evaluation across multiple operating systems. It includes a benchmark of 369 tasks covering real-world web and desktop applications, OS file I/O, and workflows across applications.

% \begin{table*}[h]
%     % \centering

%     \begin{tabular}{lc}
%         \toprule
%         \textbf{Model} & \textbf{OSWorld} \\
%         \midrule
%         \textbf{Compute-Use Agent (CUA)} \\
%         \midrule
%         OpenAI CUA & 26.0 \\
%         Claude CUA & 31.2 \\
%         OpenCUA-32B & 29.7 \\
%         \midrule
%         \textbf{GUI Agent} \\
%         \midrule
%         Qwen2.5-VL-72B & 4.4 \\
%         Augvis-72B & 10.3 \\
%         UI-TARS-72B & 22.7 \\
%         UI-TARS-72B-DPO & 24.0 \\
%         UI-TARS-1.5-7B & 24.5 \\
%         UItron-72B & 22.0\\ 
%         \bottomrule
%     \end{tabular}
%    \caption{Task Success Rates (SR) on OSWorld~\cite{OSWorld}.} 
% \end{table*}
\begin{wraptable}{r}{0.45\textwidth}
    \vspace{-5pt}
    \setlength{\tabcolsep}{4pt}
    \centering
    \small
    \begin{tabular}{lc}
        \toprule
        \textbf{Model} & \textbf{OSWorld} \\
        \midrule
        \textbf{Compute-Use Agent (CUA)} \\
        \midrule
        OpenAI CUA & 26.0 \\
        Claude CUA & 31.2 \\
        OpenCUA-32B & 29.7 \\
        \midrule
        \textbf{GUI Agent} \\
        \midrule
        Qwen2.5-VL-72B & 4.4 \\
        Augvis-72B & 10.3 \\
        UI-TARS-7B & 18.7 \\
        UI-TARS-72B & 22.7 \\
        UI-TARS-1.5-7B$^*$ & 23.3 \\
        % UI-TARS-72B-DPO & 24.0 \\
       
        UItron-72B & 24.9\\ 
        \bottomrule
    \end{tabular}
    \caption{Task Success Rates (SR) on OSWorld ~\cite{OSWorld}. We report results on their official verified environment (\emph{i.e.}, OSWorld-verified) that fix several issues. $^*$ denotes our reproduction within the same environment.}
    \label{tab:osworld_results}
    \vspace{-10pt}
\end{wraptable}

\paragraph{Baselines.} We compare our method with two types of agentic methods, namely GUI agent and computer-use agent. The GUI agent is a typical method that considers both Mobile and PC scenarios, while the compute-use agents are some recent methods that are specifically designed for PC scenario.
For GUI agent, we select several advanced baselines including Augvis-72B~\cite{xu2025aguvis}, UI-TARS-72B~\cite{qin2025ui}, UI-TARS-1.5-7B ~\cite{qin2025ui} (72B version is closed source). We also compare with Qwen2.5-VL-72B ~\cite{bai2025qwen2} to demonstrate the improvement gains via several training stages.
For compute-use agent, we select several advanced baselines including OpenAI CUA~\cite{OpenAI}, Claude CUA ~\cite{Anthropic} and OpenCUA ~\cite{wang2025opencua}.
All methods adopt the same setting of maximum length of 15 steps for fair comparison.

\paragraph{Results.} Table 8 reports the comparative results of UItron and other baseline methods.
From the results, we observe that specialized CUA agents generally outperform GUI agents, primarily due to their more singular scenarios and objectives.
We can also see that Uitron achieves competitive performance in GUI agents, with only a small gap compared to the state-of-the-art UI-Tars-1.5 method.
In addition, the experimental results also show that existing vision-language models such as Qwen25-VL suffers from poor performance, which can be greatly improved through a large amount of targeted training in GUI scenarios.

% \paragraph{AndroidWorld.}
% We use AndroidWorld~\cite{rawles2023androidinthewild} to evaluate the performance of GUI agent models as online agents on personal mobile platforms. AndroidWorld is a real-time interactive benchmark running in an Android virtual environment. It dynamically constructs tests through parameterized tasks expressed in natural language, covering 20 real Android applications and providing reward signals for 116 tasks. Each task is equipped with dedicated initialization, success-checking, and teardown logic. We report the average task success rate in our experiments.

% analysis...
% \paragraph{MobileMiniWob.}
% MobileminiWob is an instantiation of the 92 tasks from MiniWob++~\cite{zheng2023synapse} within the Androidworld environment. These tasks require agents to perform various operations through touchscreen interactions, such as clicking buttons or entering text. We utilize success-checking and action space in Androidworld to determine the task success rate.

% analysis...

\subsection{Evaluation of Chinese Scenario}

\paragraph{Evaluation Data}

We evaluate the effectiveness of our method in both offline and online environments. To support comprehensive evaluation, we constructed test data and an Android cloud environment. We manually annotate 545 trajectory steps from 109 universal tasks across several apps, and verify that these test tasks did not overlap with the training tasks.
Considering that some tasks in the online environment have some app automatic login risks and failures, we retain 86 tasks that can be completed in the online environment.

% We meticulously annotate over one million action steps from the top 100 monthly active Apps on Android platform, covering mainstream interaction scenarios such as social networking, office work, entertainment, and shopping. 
% With an average trajectory length of 5 steps, our dataset comprises a substantial proportion of multi-round human-agent interactions that enable user's preference elicitation when necessary. This design promotes proactive ambiguity identification in initial instructions, requiring a deeper comprehension on GUI-based tasks.
% Based on this, we constructed an offline evaluation dataset to assess the capabilities of GUI agents in Chinese App scenarios, evaluating the performance of different models based on three metrics: task grounding, single-step success rate, and task completion rate. 
% To evaluate the realistic interaction performance of GUI agent in real-world applications, we also build an online Android mobile environment for evaluation.

\paragraph{Evaluation Metrics}

We design different evaluation metrics for GUI agents in offline and online environments. For evaluation in offline environment, in which each predicted action corresponding to a ground-truth action, we directly calculate accuracy to evaluate the single-step success rate (\emph{i.e.}, Step SR) and task success rate (\emph{i.e.}, Task SR). 
% When Task SR is equal to 1, it means that all Step SR under this task are 1.
A task is deemed successful when all execution steps exactly align with the ground-truth action sequence.
For evaluation in online environment, the GUI agent freely explores all possible actions according to finish the task without any ground-truth action. Therefore, we evaluate whether the task is completed accurately based on the complete execution trajectory.
We leverage an advanced visual-language model (\emph{i.e.}, GPT-4o \cite{hurst2024gpt}) to determine whether the task is completed. The result is 1 for completion and 0 for incomplete.

% \begin{table*}[h]
% \centering
% \begin{adjustbox}{max width=1.0\textwidth}
% \begin{tabular}{lcc}
% \toprule
% \textbf{Method} & \textbf{Step SR} & \textbf{Task SR}  \\
% \midrule
% UI-TARS-1.5-7B & 77.4 & 29.3 \\
% UI-TARS-7B & 75.1 & 22.4 \\
% UI-TARS-72B & \\
% \midrule
% UItron-7B & 82.7 & 40.5\\
% UItron-72B & \textbf{84.1} & \textbf{47.4}  \\
% \bottomrule
% \end{tabular}
% \end{adjustbox}
% \caption{Offline evaluation results on our Chinese GUI scenarios.}
% \label{tab:cn_offline}
% \end{table*}
\begin{wraptable}{r}{0.45\textwidth}
    \vspace{-5pt}
    \setlength{\tabcolsep}{4pt}
    \centering
    \small
    \begin{tabular}{lcc}
        \toprule
        \textbf{Method} & \textbf{Step SR} & \textbf{Task SR}  \\
        \midrule
        \multicolumn{2}{l}{\textbf{Offline Environment}} \\
        \midrule
        UI-TARS-7B & 75.1 & 22.4 \\
        UI-TARS-72B & 80.5 & 32.8 \\
        UI-TARS-1.5-7B & 77.4 & 29.3\\
        % \midrule
        UItron-7B & 82.7 & 40.5\\
        UItron-72B & 84.1 & 47.4\\
        \bottomrule
    \end{tabular}
    \caption{
       Offline evaluation results on top-tier Chinese mobile Apps. 
    }
    \label{tab:cn_offline}
    \vspace{-10pt}
\end{wraptable}

\paragraph{Offline Results}

Table~\ref{tab:cn_offline} reports the Step SR and Task SR results of UItron and baseline methods.
From the results, we can see that both UItron-7B and UItron-72B significantly outperform the baseline methods in all evaluation metrics, demonstrating their superiority in Chinese scenarios. Interestingly, for the Step SR and Task SR indicators, the results indicate that they have a positive correlation, but the difference in Task SR is significantly larger, which is probably because Task SR reflects the more rigorous accumulation of Step SR.
Therefore, we can see from the results that different methods are relatively close in terms of Step SR, but have significant differences in terms of Task SR.
The advanced performance of UItron primarily stems from learning page organization and interaction logic through extensive data from Chinese scenarios, which exhibit significant differences compared to traditional English contexts.

\begin{wraptable}{r}{0.45\textwidth}
    \vspace{-5pt}
    \setlength{\tabcolsep}{12pt}
    \centering
    \small
    \begin{tabular}{lc}
        \toprule
        \textbf{Method} & \textbf{Task SR}  \\
        \midrule
        \multicolumn{2}{l}{\textbf{Online Environment}} \\
        \midrule
        % UI-TARS-7B & & \\
        UI-TARS-1.5-7B & 38.9 \\
        % UI-TARS-72B & 49.4 \\
        % \midrule
        UItron-7B & 44.4 \\
        UItron-72B & 54.1  \\
        \bottomrule
    \end{tabular}
    \caption{
       Online evaluation results on top-tier Chinese mobile Apps. 
       % Due to inherent heterogeneity between our physical devices, the execution environment does not keep exactly the same throughout the entire evaluation process, and thus the number of trails varied  during model evaluation.
    }
    \label{tab:cn_online}
    \vspace{-10pt}
\end{wraptable}

\paragraph{Online Results}
Table~\ref{tab:cn_online} reports the Task SR results of UItron and baseline methods.
The results indicate that UItron outperforms the baseline model with a significant performance advantage, verifying its better interaction and exploration capabilities in online environments.
Another noteworthy phenomenon is that the online evaluation results of the same model consistently surpass its offline evaluation results, a trend often overlooked in previous research due to the lack of comparable offline and online tasks.
The explanation for this phenomenon lies in the nature of the online environment, which offers GUI agents ample space to explore and recover from errors with relaxed constraints. In this setting, certain erroneous steps can be rectified by returning to the original step. Conversely, in offline evaluations, any failed step inevitably results in task failure.

\section{Conclusion and Future Work}
This paper presents UItron, a pioneering open-source foundational model designed to enhance the capabilities of GUI agents in executing complex tasks across digital environments such as PCs and Mobile devices. UItron conduct sufficient investigation of data engineering and interactive infrastructure to handle the scarcity of annotated trajectory data. It systematically compares various data strategies to improve the training effectiveness.
UItron adopts a typical training paradigm of GUI grounding and planning, and then develops a curriculum reinforcement learning method that improves complex reasoning and exploration in the online environment. In particular, UItron also emphasizes the importance of Chinese interaction capabilities in practical GUI agent deployment.
Through comprehensive annotation of over one million action steps from leading Chinese apps, UItron achieves superior results in realistic offline and online evaluation frameworks, bringing GUI agents closer to practical deployment. 
Experimental results demonstrate that UItron achieves superior performance in benchmarks of GUI perception, task localization and planning, as well as a significant advance in Chinese application scenarios.

In summary, UItron offers an open-source foundation that facilitates the future development of GUI agents. In our future work, we will investigate the intrinsic thinking patterns underlying the interpretive action behaviors of GUI agents, as we have observed frequent ambiguities and inconsistencies between thinking and action outputs in our method. Furthermore, we will study multi-agent collaboration strategies to systematically explore capabilities such as reflection, backtrace, and correction, considering that current single-agent methods often struggle to simultaneously handle both visual and textual aspects. Lastly, we plan to build the unified agentic infrastructure and reinforcement learning environment that integrate capabilities such as coding, tool-use and function-call, spanning both the 2D digital world and the 3D physical realm.

\begin{table*}[h]
\centering

\begin{adjustbox}{max width=1\textwidth}
\begin{tabular}{@{}lccccccccc@{}}
\toprule
    \textbf{Environment} & \textbf{Action Space} \\ 
\midrule
    \multirow{7}{*}{Web} & \texttt{\{"name": "computer\_use", "arguments": \{"action": "key", "keys": ["ctrl", "a"]\}\}} \\
                         & \texttt{\{"name": "computer\_use", "arguments": \{"action": "left\_click", "coordinate": [x, y]\}\}}  \\
                         & \texttt{\{"name": "computer\_use", "arguments": \{"action": "type", "text": "text"\}\}} \\ 
                         & \texttt{\{"name": "computer\_use", "arguments": \{"action": "answer", "text": "text"\}\}} \\ 
                         & \texttt{\{"name": "computer\_use", "arguments": \{"action": "terminate", "status": ["success"]\}\}} \\ 
                         & \texttt{\{"name": "computer\_use", "arguments": \{"action": "wait", "time": "time"\}\}} \\ 
                         & \texttt{(Total 15 action types...)} \\
\midrule
    \multirow{8}{*}{Mobile} & \texttt{\{"name": "mobile\_use", "arguments": \{"action": "system\_button", "button": "enter"\}\}} \\
                           & \texttt{\{"name": "mobile\_use", "arguments": \{"action": "click", "coordinate": [x, y]\}\}} \\ 
                           & \texttt{\{"name": "mobile\_use", "arguments": \{"action": "type", "text": "text"\}\}} \\ 
                           & \texttt{\{"name": "mobile\_use", "arguments": \{"action": "swipe", "coordinate": [x, y], "coordinate2": [x, y]\}\}} \\ 
                           & \texttt{\{"name": "mobile\_use", "arguments": \{"action": "type", "text": "text"\}\}} \\ 
                           & \texttt{\{"name": "mobile\_use", "arguments": \{"action": "status", "button": "success"\}\}} \\ 
                           & \texttt{\{"name": "mobile\_use", "arguments": \{"action": "wait", "time": "time"\}\}} \\ 
                           & \texttt{(Total 11 action types...)} \\
                           
\bottomrule
\end{tabular}
\end{adjustbox}
\caption{Action space specifications for Web and Mobile environments}
\label{tab:action_space}
\end{table*}

% \begin{ack}
% Use unnumbered first level headings for the acknowledgments. All acknowledgments

% Do {\bf not} include this section in the anonymized submission, only in the final paper. You can use the \texttt{ack} environment provided in the style file to automatically hide this section in the anonymized submission.
% \end{ack}

\bibliographystyle{plain}
\bibliography{main}

% \section*{References}

% References follow the acknowledgments in the camera-ready paper. Use unnumbered first-level heading for
% the references. Any choice of citation style is acceptable as long as you are
% consistent. It is permissible to reduce the font size to \verb+small+ (9 point)
% when listing the references.
% Note that the Reference section does not count towards the page limit.
% \medskip

% {
% \small

% [1] Alexander, J.A.\ \& Mozer, M.C.\ (1995) Template-based algorithms for
% connectionist rule extraction. In G.\ Tesauro, D.S.\ Touretzky and T.K.\ Leen
% (eds.), {\it Advances in Neural Information Processing Systems 7},
% pp.\ 609--616. Cambridge, MA: MIT Press.

% [2] Bower, J.M.\ \& Beeman, D.\ (1995) {\it The Book of GENESIS: Exploring
%   Realistic Neural Models with the GEneral NEural SImulation System.}  New York:
% TELOS/Springer--Verlag.

% [3] Hasselmo, M.E., Schnell, E.\ \& Barkai, E.\ (1995) Dynamics of learning and
% recall at excitatory recurrent synapses and cholinergic modulation in rat
% hippocampal region CA3. {\it Journal of Neuroscience} {\bf 15}(7):5249-5262.
% }

%%%%%%%%%%%%%%%%%%%%%%%%%%%%%%%%%%%%%%%%%%%%%%%%%%%%%%%%%%%%

% \appendix

% \section{Technical Appendices and Supplementary Material}
% Technical appendices with additional results, figures, graphs and proofs may be submitted with the paper submission before the full submission deadline (see above), or as a separate PDF in the ZIP file below before the supplementary material deadline. There is no page limit for the technical appendices.

%%%%%%%%%%%%%%%%%%%%%%%%%%%%%%%%%%%%%%%%%%%%%%%%%%%%%%%%%%%%

\newpage

\end{document}